\documentclass[11pt]{article}

\usepackage[final]{acl}

\usepackage{times}
\usepackage{latexsym}
\usepackage{amsmath,amsfonts}
\usepackage[T1]{fontenc}
\usepackage{algorithm}
\usepackage{algpseudocodex}
\usepackage{booktabs} 
\usepackage[utf8]{inputenc}

\usepackage{microtype}
\usepackage{hyperref}
\usepackage{inconsolata}

\usepackage{graphicx}
\usepackage{xspace}
\usepackage[hide]{todo}
\usepackage{xcolor}
\usepackage{alltt}
\definecolor{navy}{RGB}{0,0,139}

\newcommand{\sent}{\ensuremath{\mathbf{w}}}
\newcommand{\indic}[1]{\ensuremath{\mathbb{I}[#1]}}

\newcommand{\genericname}{ILP4LID\xspace{}}
\newcommand{\fasttext}{\mbox{FastText}\xspace}
\newcommand{\wordlid}{\mbox{LiteLID}\xspace}
\newcommand{\masklid}{MaskLID\xspace{}}
\newcommand{\glotlid}{GlotLID\xspace{}}
\newcommand{\alphapar}{M}
\newcommand{\interval}[2]{\ensuremath{[#1\hspace{-0.5ex}:\hspace{-0.5ex}#2]}}
\usepackage{fontawesome}
\usepackage{enumitem}
\setlist{topsep=0pt}
\title{More than one language! Extending Language Identification to Code-Switched Texts}
\title{More than one language! From Language Identification to LanguageS Identification}
\title{OneOrMoreLID in code-switched settings: \textit{Plus d'une langue}}
\title{PolyLID for code-switched utterances: \textit{reconnaître plus d'une langue}}
\title{\textit{Plus d'une langue !}\thanks{French for ``More than one language!''} Language Identification for Code-Switched Utterances}
\title{MixLID: Improving Language Identification for Code-Switched Utterances with Integer Linear Programming}
\title{Improving Language Identification for Code-Switched Utterances with Integer Linear Programming}

\author{Joanna Rado\l{l}la \\
  Affiliation / Address line 1 \\
  Affiliation / Address line 2 \\
  Affiliation / Address line 3 \\
  \texttt{email@domain} \\ \And
  Josep Maria Crego \\
  Affiliation / Address line 1 \\
  Affiliation / Address line 2 \\
  Affiliation / Address line 3 \\
  \texttt{email@domain} \\ \And
  François Yvon \\
  Affiliation / Address line 1 \\
  Affiliation / Address line 2 \\
  Affiliation / Address line 3 \\
  \texttt{yvon@isir.umpc.fr} \\
}

\author{
  \textbf{Joanna Rado\l{}a\textsuperscript{1,2}}\hspace{0.5cm}
  \textbf{Josep Maria Crego\textsuperscript{2}}\hspace{0.5cm}
  \textbf{François Yvon\textsuperscript{1}}
\\
  \textsuperscript{1}Sorbonne Université, CNRS, ISIR, Paris, France\\
  \textsuperscript{2}SYSTRAN by ChapsVision, Paris, France
\\
  \small{
    \textbf{Correspondence:} \href{mailto:radola@isir.upmc.fr}{radola@isir.upmc.fr}
  }
}

\begin{document}
\maketitle

\begin{abstract}\done\todo{rewrite abstract}
  Automatic identification of code-switched (CS) utterances remains a challenge for language identification (LID) systems, causing such texts to be underrepresented in the training data of Large Language Models. In this paper, we revisit \masklid{}, a state-of-the art approach for CS identification, which requires no training and detects arbitrary language combinations. We make three main contributions: (a) we reveal, and address, a major issue of \masklid{}: its overreliance on word-level language association scores; (b) we reformulate the underlying optimization algorithm as an Integer Linear Program, enabling us to experiment with a large set of clear and interpretable constraints; (c) each of these improvements vastly improves the baseline system, as we illustrate in experiments involving 10~diverse languages, where we observe a strong boost in performance on CS benchmarks.
  We release our code and data for reproducibility.
  
\centering  \faicon{github} \href{https://github.com/jradola/ILP4LID-dev}{github.com/jradola/ILP4LID}
\end{abstract}

\todo{colorblind clashes with aclstyles}
\section{Introduction}
\label{sec:introduction}
The development of multilingual language technologies (LTs), illustrated by the rise of multilingual models trained on webscale corpora \citep{imanigooghari-etal-2023-glot500,kudugunta-etal-2023-madlad,ustun-etal-2024-aya}, requires effective multilingual natural language processing (NLP) tools, capable to handle as many languages as possible. Among these, language identification (LID) tools stand out as they are usually used in early stages of the data collection and filtering \citep{abadji-etal-2021-ungoliant,kargaran-etal-2024-glotcc,penedo-etal-2025-fineweb}: high-precision LIDs are especially critical for low-resource languages, which can easily get mixed with dominant languages, thereby compromising the quality of the resulting LTs \citep{kreutzer-etal-2022-quality}.    

In its basic form, Language Identification (LID) is framed as a supervised text classification task, where the language code of the text is the target label to be predicted. Modern LIDs detect several hundreds languages, with recent work claiming to cover over a thousand~\citep{brown-2014-non,dunn-2020-mapping,costa-jussa-etal-2024-scaling,adebara-etal-2022-afrolid,jauhiainen-etal-2022-heli,burchell-etal-2023-open,kargaran-etal-2023-glotlid,kargaran-etal-2024-glotcc}. Many of these rest on a \fasttext{} backbone \citep{bojanowski-etal-2017-enriching}, which provides an efficient implementation for text classification at scale and also ships with a pretrained LID model recognizing 176~languages.\footnote{\url{https://fasttext.cc/docs/en/language-identification.html}} 

Challenges of contemporary LIDs are reviewed in e.g., \citep{jauhiainen-etal-2018-automatic,caswell-etal-2020-language, burchell-etal-2024-code,goot-2025-identifying}. They include the discrimination of typologically related languages and the computation of well-calibrated posterior probabilities.
LIDs also often struggle with short texts, especially when written in a non-standard orthographies, as social media posts. A last challenge, that we address in this work, is their inability to handle intra-sentential \textsl{code-switching} (CS), where several languages simultaneously occur in the same sentence. This phenomenon is particularly widespread in informal communication contexts \citep{dogruoz-etal-2021-survey}. 

Our starting point is \masklid{} \citep{kargaran-etal-2024-masklid}, which turns generic LIDs into CS LIDs thanks to a simple post-processing layer. This approach is conceptually simple, requires no training and enables to \emph{recognize arbitrary language combinations}. It thus provides us with a strong baseline, with many desirable properties.

We make three main contributions: (a) we reveal a major issue of \masklid{}: its over-reliance on inaccurate word-level language association scores that we address using an improved underlying LID; (b) we reformulate the language assignment algorithm as an Integer Linear Program, enabling us to experiment with a large set of clear and interpretable constraints on CS utterances; (c) we observe, in experiments involving 10 languages, both low and high resource, that each of these improvements vastly improve the baseline system, with boosts in accuracy up to +100\% on several CS benchmarks.

\section{Background \label{sec:methods}}

\subsection{\fasttext-based LIDs}\done\todo{Make this 10 lines shorter}
\label{sec:fasttextlids}

Following \citep{kargaran-etal-2024-masklid}, we rely on LIDs based on the \fasttext{}\done\todo{add macro for consistency} \citep{bojanowski-etal-2017-enriching} architecture. \fasttext is an open-source text-classification framework that can be deployed at scale at a reasonable computational cost. It still constitutes the backbone of several state-of-the-art LIDs \citep{suarez-etal-2026-commonlid}. A \fasttext model is a multinomial logistic classifier that represents its input text as set of word embeddings\footnote{We define words as tokens computed by the \fasttext tokenizer, which considers whitespaces as token boundaries.}  (vectors in $\mathbb{R}^d$) extracted from the text.
The embedding $\mathbf{e}(w)$ of word $w$ is a summation of the embeddings $\mathbf{g}(u)$ of n-grams $u$ occurring in $w$.\footnote{Up to a maximum value of $n$, set to $6$ by default.} Sufficiently frequent words also have a dedicated embedding $\mathbf{f}(w)$. For those words, $\mathbf{e}(w) = \mathbf{f}(w) + \sum_{g \in w} \mathbf{g}(g)$.

LID systems relying on \fasttext{} compute the posterior probability of a language $l \in [1:L]$ by applying the $\operatorname{softmax}$ function to sentence-level representations $\mathbf{e}(\sent) = \frac{1}{T} \sum_{t=1}^{T} \mathbf{e}(w_t)$:
\begin{align*}\label{eq:predict}
  P(l|\sent=w_1 \dots w_T) = \frac{\exp( \mathbf{b}_{l} \cdot \mathbf{e}(\sent))}{\sum_{l'=1}^{L} \exp(\mathbf{b}_{l'} \cdot \mathbf{e}(\sent))},
\end{align*}
where $\mathbf{b}_l$ is the embedding of language $l$. Logits can also be computed separately for each word, as they simply add up in the sentence representation. We thus assume a base LID computing a $L \times T$ tensor \(\mathbf{C}(\sent)\) for each input \sent, where cell \(\mathbf{C}_{l,t}(\sent)\) stores the logits for language $l$ and word-level feature $\mathbf{e}(w_t)$ computed as the dot product:
\begin{equation}\label{eq:matrix_v}
  \mathbf{C}_{l,t}(\sent) = \mathbf{b}_{l} \cdot \mathbf{e}(w_t).
\end{equation}

\subsection{\masklid{}: Iterative Masking \label{ssec:masklid}}
\masklid{} \citep{kargaran-etal-2024-masklid} rests on a simple assumption: in a CS sentence containing two languages, L1 (dominant) and L2 (embedded), the corresponding fragments in L1 and L2 should be mapped to their respective language by any LID system. This idea is implemented in the following procedure for a text $\sent=w_1 \dots w_T$:
\begin{enumerate}[parsep=0pt]
\item predict language $l^*$ for $\sent$ using any LID
\item for each word $w_t, t=1 \dots T$:
  \begin{enumerate}
  \item run LID on $w_t$, compute the set of $\alphapar$ most likely languages $\mathcal{L}_{\alphapar}(t)$
  \item if $l^* \in \mathcal{L}_{\alphapar}(t)$, assign $w_t$ to $l^*$ and
    remove $w_t$\footnote{The exact implementation is slightly more complex, as $\alphapar$ \emph{may vary} depending on the total predicted probability of $l^*$. See pseudo-code and discussion in Appendix~\ref{sec:allmasklid}.} from \sent
  \end{enumerate}
\item if $\operatorname{length}(\sent) \ge{} \tau$ goto 1, else terminate
\end{enumerate}

\paragraph{}The key insight of \masklid{}  is step~(2.b), which assigns $w_t$ to $l^*$ whenever $l^*$ is one of the $\alphapar$ most likely language for $w_t$. This ensures that these words will not be available in later iterations, where they could cause the prediction of languages closely resembling $l^*$. Assume, for instance, that some words in a monolingual German sentence would have Dutch, a related language, as their most likely language, while German is only ranked second or third. Step~(2.b) assigns these words to German, preventing their assignment to Dutch in a subsequent round. The algorithm stops in (3) when the remaining part of \sent{} is too short to be reliably classified, as controlled by the length parameter $\tau$ - potentially leaving some words unassigned. The procedure returns the set of languages iteratively identified in step~(1). 

This procedure is illustrated in the following execution trace borrowed from \citep{kargaran-etal-2024-masklid}. In the first pass, $l^*$ is predicted to be Turkish, causing all \underline{underlined words} to be removed from further analysis, leaving the fragments ``deadline crash walking I heard it at study" for the second iteration. As this text is sufficiently long, English is identified as the second language. Once these words are removed, the algorithm terminates and returns $\{$tur, eng$\}$.

\begin{quote}
  \underline{ya} \textcolor{red}{deadline} \underline{gelmişti çok büyük bir}
  \textcolor{red}{crash} \underline{olmuş arkadaşlarla}
  \textcolor{red}{walking} \underline{yaparken} \textcolor{red}{I heard it at}
  \underline{boğaziçi sesli} \textcolor{red}{study}
\end{quote}

Compared to most alternatives reviewed in \textsection\ref{sec:related-work}, the main benefits of \masklid{} are (a) CS LID can be performed for any language combination that is recognized by the LID: using large coverage models, such as \glotlid{} (\fasttext{}-based), recognizing 2000+ languages, \masklid{} readily identifies more than 4M combinations of languages;  (b)~\masklid{} can work with any LID, as long as it outputs posterior language probabilities given an input text (a sentence or a word). In the experiments of \citep{kargaran-etal-2024-masklid}, 2 backbone LIDs and 4~language pairs are considered, and \masklid{} is found to vastly outperform the approach of \citep{burchell-etal-2024-code} on the code-switched subset of the test data.

\subsection{\masklid{}: a reanalysis \label{ssec:reanalysis}}
\done\todo{Explain the optimization that is performed}
As illustrated above, \masklid{} computes \emph{sentence-level probabilities and assignments}; internally, however, it mostly relies on \emph{word-level associations} (step 2.a), which identify words to be iteratively masked. Such associations, when computed by a classifier to perform sentence-level LID, are not reliable. We start in \textsection\ref{sec:word-level-lid} by documenting and addressing this shortcoming with the development of an improved backbone LID model.

Sentence-level assignments do not just depend on word-level scores, they also take into account global scores (step~1), as well as length constraints. 
\masklid{} can thus be understood as computing maximally likely word associations, subject to global constraints. It solves this problem in a greedy fashion, iteratively computing optimal mappings for diminishing input spans. We instead formalize in \textsection{}\ref{sec:ilpmodels} this combinatorial problem as an Integer Linear Program (ILP), hoping to reach  \textbf{better global solutions}. By further extending the basic model with additional constraints, we create a versatile tool for code-switched language detection.
 
\section{Improving \masklid{} assignments with better word-level scores \label{sec:word-level-lid}}
\todo{Write a small para to organize this section}

In this section, we report preliminary experiments highlighting a mismatch between \masklid{}, which heavily relies on word-level language scores, and standard LIDs, which are typically trained at the sentence level, and deliver poor word-level predictions. We then document our improved LID, which can both reliably identify languages at the sentence level, and serve as an effective word-level language predictor for \masklid. We begin with a presentation of the datasets and metrics.

\subsection{Datasets and evaluation metrics}

\paragraph{Code-switched datasets} We use the following CS corpora from the literature: Turkish-English \citep{yirmibesoglu-eryigit-2018-detecting}, Turkish-German \citep{cetinoglu-2016-turkish}, Basque-Spanish \citep{aguirre-etal-2022-basco,heredia-etal-2025-euskanolds}, Hindi-, Nepali-\done\todo{Missing references - all three come from LinCE}\footnote{Hindi and Nepali texts are romanized in this corpus, sourced from social media.}, Spanish-English \citep{aguilar-etal-2020-lince},  Indonesian-English \citep{barik-etal-2019-normalization} and Wolof-French \citep{gauthier-etal-2024-kallaama}. 
As these datasets were independently sourced from social media or from transcripts of conversational corpora, we had to reconcile heterogeneous annotation schemes. For instance, some datasets distinguish between Named Entities, or label some tokens as language-agnostic, see discussion in \citep{sterner-2024-multilingual}.\todo{rewrite: after preprocessing (appendix \ref{}), each CS sentence is labeled with a set of two labels.} For corpora tagged at the word level, we only keep examples that include words tagged as L1 or L2.\footnote{For example, we exclude sentences including only Turkish and \emph{Mixed} labels; or only Indonesian and \emph{Unknown}.} Emojis, URLs, @usernames, quotation marks, tokens starting with "\%" which indicate filler words (e.g. ``\%hmm'') and language tags have been removed. We filter out examples that are shorter than 20 and longer than 200 characters excluding whitespaces. Detailed statistics regarding the data are in Table~\ref{tab:cs-statistics}, which highlights the heterogeneity of these benchmarks, e.g., regarding the length of L2 spans. 


\begin{table*}[htbp]
    \centering
    \resizebox{\textwidth}{!}{%
      \begin{tabular}{lcccccccc}
    \hline
                          & tur-eng               & eus-spa   & hin-eng               & npi-eng                 & tur-deu               & spa-eng               & ind-eng               & fra-wol  \\ \hline
    \# sent in dev / test & 100 / 256             & 100 / 965 & 100 / 1928            & 100 / 6959              & 100 / 1143            & 0 / 8269              & 0 / 627               & 0 / 4386 \\
    mean \# char/sent     & 84                    & 79        & 72                    & 56                      & 75                    & 57                    & 103                   & 73       \\
    mean len L1           & 57.7\small[$\pm$29.4] & -         & 47.0\small[$\pm$32.0] & 36.4\small{[$\pm$18.4]} & 50.6\small[$\pm$27.9] & 41.1\small[$\pm$20.7] & 64.7\small[$\pm$36.4] & -        \\
    mean len L2           & 19.0\small[$\pm$11.9] & -         & 13.1\small[$\pm$11.4] & 12.4\small[$\pm$8.6]    & 21.4\small[$\pm$14.3] & 11.2\small[$\pm$8.3]  & 24.0\small[$\pm$17.0] & -        \\
    mean \# switches      & 2.8                   & -         & 2.9                   & 2.8                     & 1.9                   & 2.9                   & 2.8                   & -        \\
    type                  & sm                    & conv+sm   & sm                    & sm                      & conv                  & sm                    & sm                    & conv     \\
    \hline
\end{tabular}
    }
    \caption{Statistics of the CS datasets used the experiments. The uncertainties reported are standard deviations. The eus-spa and fra-wol datasets do not contain the word-level labels necessary to compute relevant CS-specific statistics. sm=social media corpus, conv=conversational corpus.}
    \label{tab:cs-statistics}
\end{table*}\done\todo{make the presentation more compact - include the type ? - change the order if we have time}

To create the devset, we randomly select 100 sentences from five CS corpora and also include 500~monolingual sentences in these same languages.\footnote{eus, tur, spa: 72 sentences / lang., hin, npi, eng, deu: 71.} The remaining data is reserved for testing.\done\todo{New dev splits}

\paragraph{Monolingual evaluations} A good CS LID should also reliably detect non-CS cases. Our evaluation data thus balances monolingual and code-switched examples. Some CS datasets, e.g. those included in LinCE \citep{aguilar-etal-2020-lince}, 
include a monolingual subset. To balance out those that do not, we include monolingual sentences sourced from Flores+ \citep{costa-jussa-etal-2024-scaling} in the dataset. We take 500 shortest sentences from every language out of the 10 appearing in our CS pairs. Note that these examples are considerably cleaner than the other monolingual data.

Hindi and Nepali pose specific problems, as they are written in Latin scripts in the CS corpora, but use Devanagari in Flores+. We automatically romanize the Flores texts for these two languages using the \texttt{uroman} Python module\footnote{\url{https://github.com/isi-nlp/uroman}} and include romanized monolingual sentences in the test set.

\subsection{Metrics}
Our framing of CS LID associates each input sequence with a set of languages. Like \citep{kargaran-etal-2024-masklid}, our primary metric is the \textsl{exact match} (EM), which computes the proportion of samples for which the output set of languages exactly matches the reference. As we predict sets, we also compute precision, recall, and F1, and average those per language over input samples. Aggregated results are always macro-averages across languages or language pairs.

\subsection{\glotlid{} scores cannot always be trusted  \label{ssec:issues-confidence}}

In our early experiments, we ran \masklid{} combined with \glotlid{} and observed that \glotlid{} was sometimes predicting unexpected labels for short segments. For instance, ``\textsl{i the same lady how u doing}'' was labeled as Norwegian (nno), which was also the most likely language for \textsl{same} and \textsl{lady}.

\begin{figure}[h]
    \centering
    \includegraphics[width=0.49\linewidth]{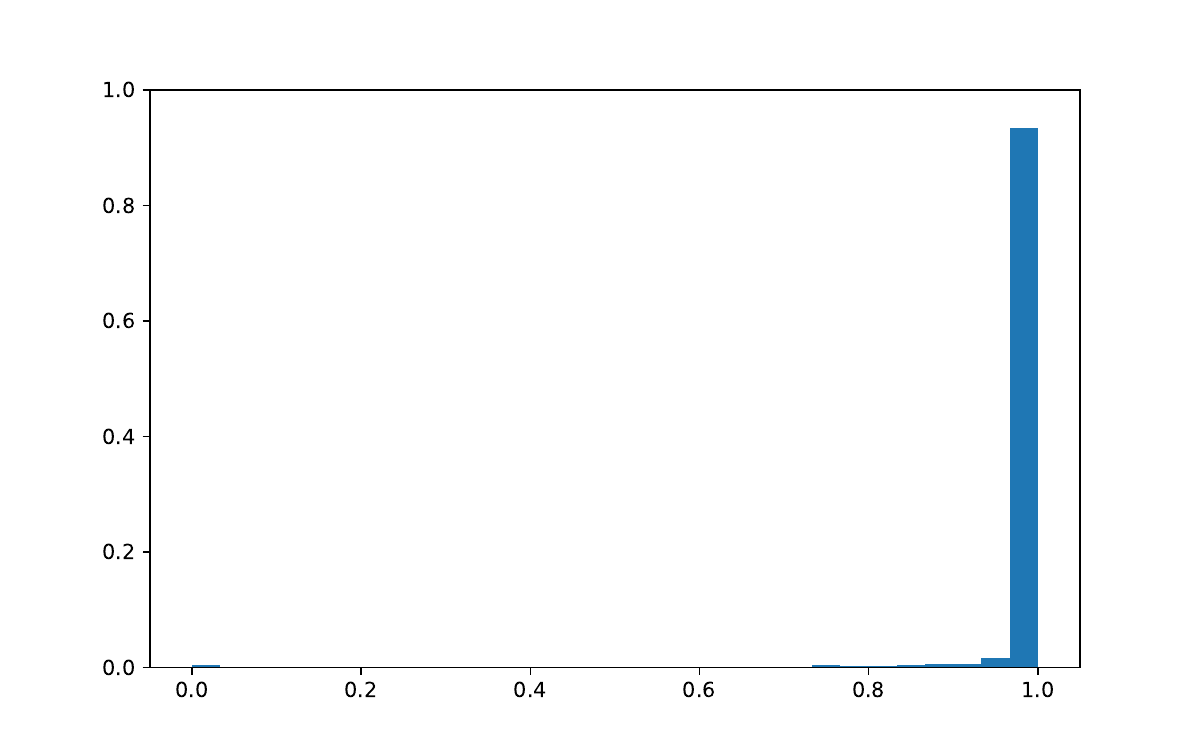}
    \includegraphics[width=0.49\linewidth]{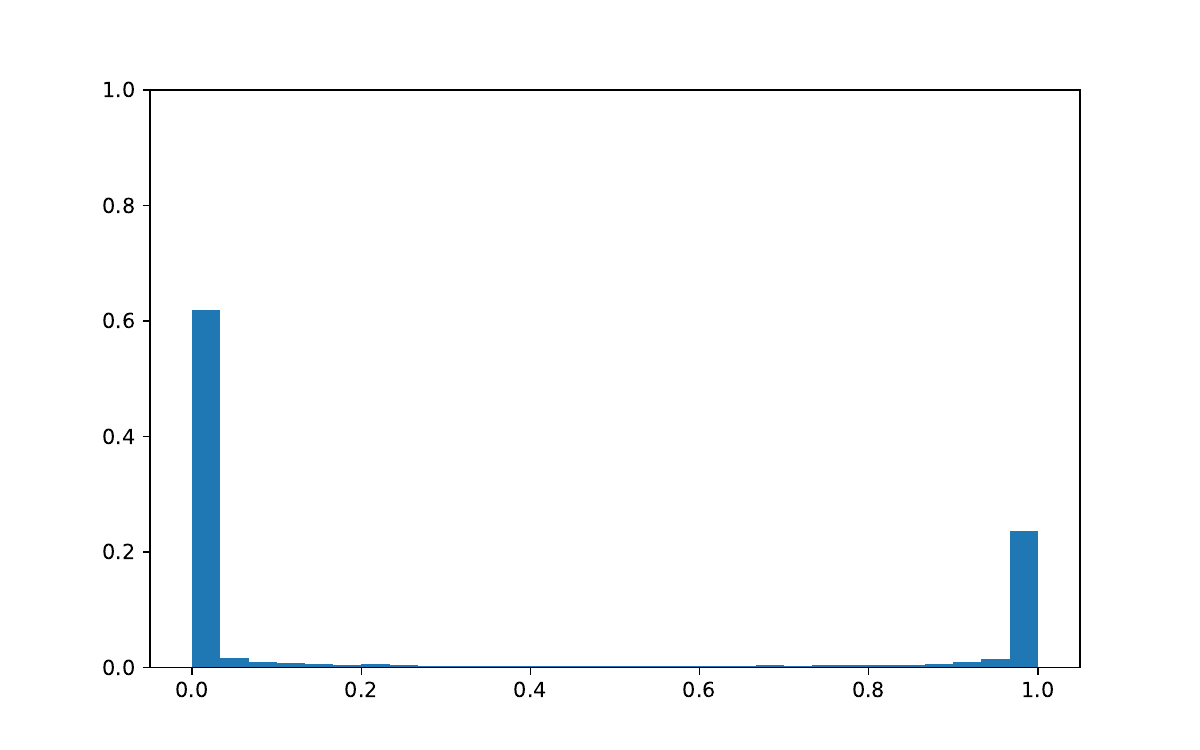 }
    \includegraphics[width=0.49\linewidth]{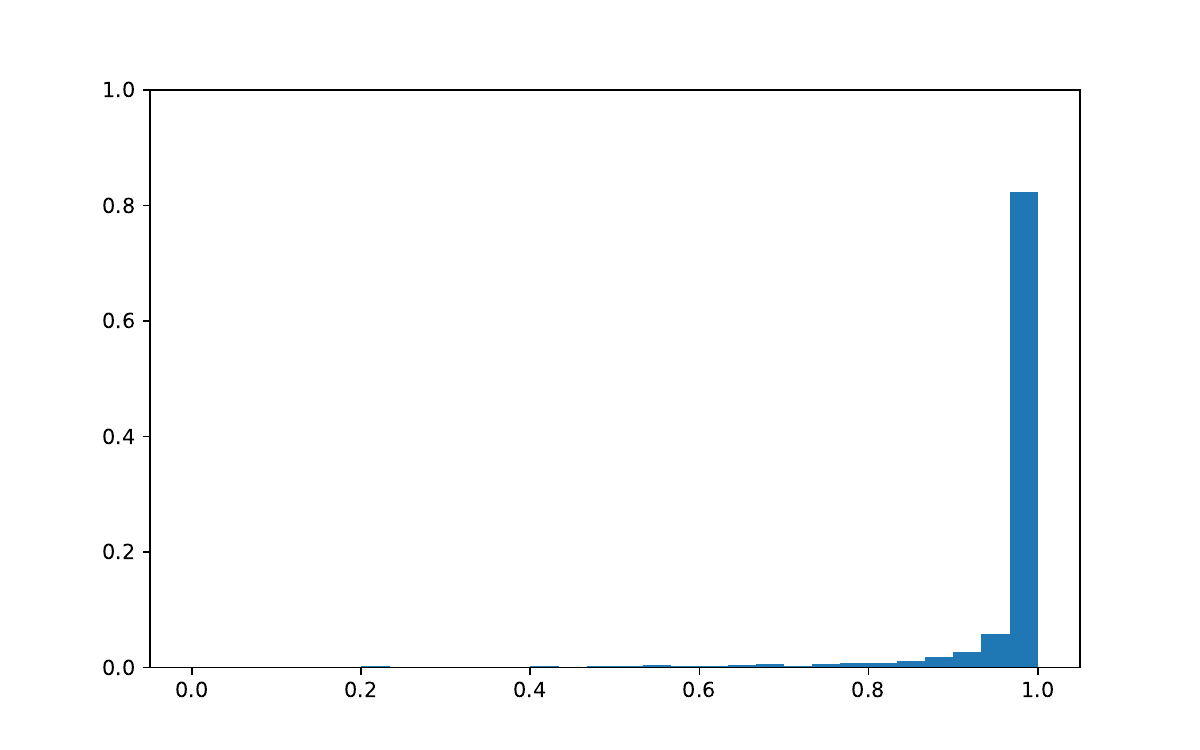}
    \includegraphics[width=0.49\linewidth]{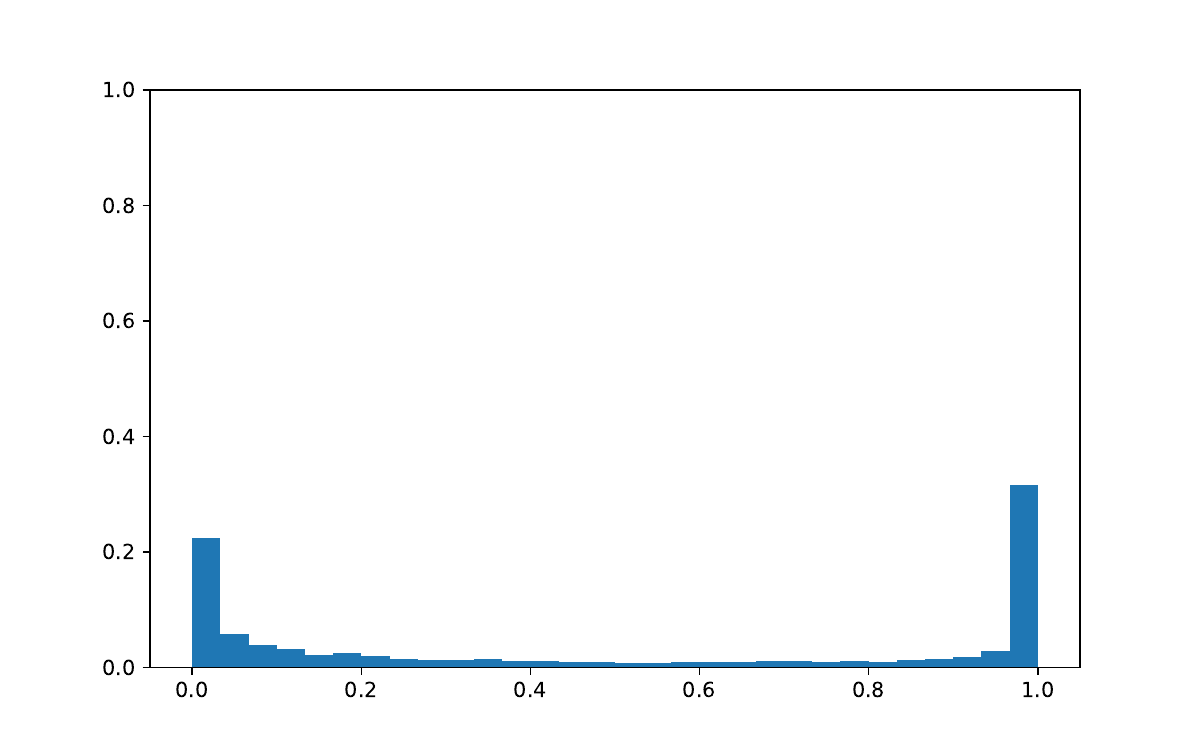}
    \caption{Top: GlotLID distribution of probabilities assigned to the gold label on a subset of FLORES sentences (left) and words (right). Mean score for a sentence: 0.98. Mean score for a word: 0.31. Bottom: Same figure for \wordlid{}-v2. Mean score for a sentence: 0.96. Mean score for a word: 0.50. 
    }
    \label{fig:glotsentword}
\end{figure}

To visualize the general brittleness of word-level predictions, we plot on Figure~\ref{fig:glotsentword} the distribution of posterior probability of the reference language at the sentence (left) and word-levels (right). While almost all sentence-level scores are close to 1.0, we see that at the level of words, a vast majority of gold labels have a near 0 probability. This highlights the fact that one key assumption of \masklid, i.e that the underlying word-level LID scores can be trusted to select masked words, does not hold. 

\done\todo{To make our point we need to retrain GlotLID at the sentence level with 127 labels. Otherwise we cannot really conclude.}

\subsection{\wordlid{}: Enriching \glotlid{} with word-level training examples}

To precisely measure the impact of this issue on \masklid{}'s performance, we trained several new \fasttext{}-based LIDs on a set of 125~labels, corresponding to the languages written in Latin script in \fasttext-LID.\footnote{This reduced label set is also adopted by \cite{kargaran-etal-2024-masklid}.} These models are denoted as \wordlid{}.
Our baseline model (\wordlid{}-v0) reproduces the training settings of \glotlid{}, using the Glot-C corpus of \citep{kargaran-etal-2024-glotcc}; we then consider multiple variants where the training data comprises a mixture of complete sentences, short segments and isolated words. Details on the experiments with data augmentation schemes are in \ref{sec:training-versions}. In the remainder, we select to work with \wordlid{}-v2\done\todo{modelname}, which displays good performance at the sentence and word levels. This model was trained on full sentences and single-word training instances, obtained from splitting full sentences.

The difference between this model and \glotlid{} is visible in Figure~\ref{fig:glotsentword}, where in the bottom we see a clear improvement of average word-level scores, and a sharp decrease of the proportion of cases where the correct label has a near-zero probability.

\begin{figure}
    \centering
    \includegraphics[width=0.75\linewidth]{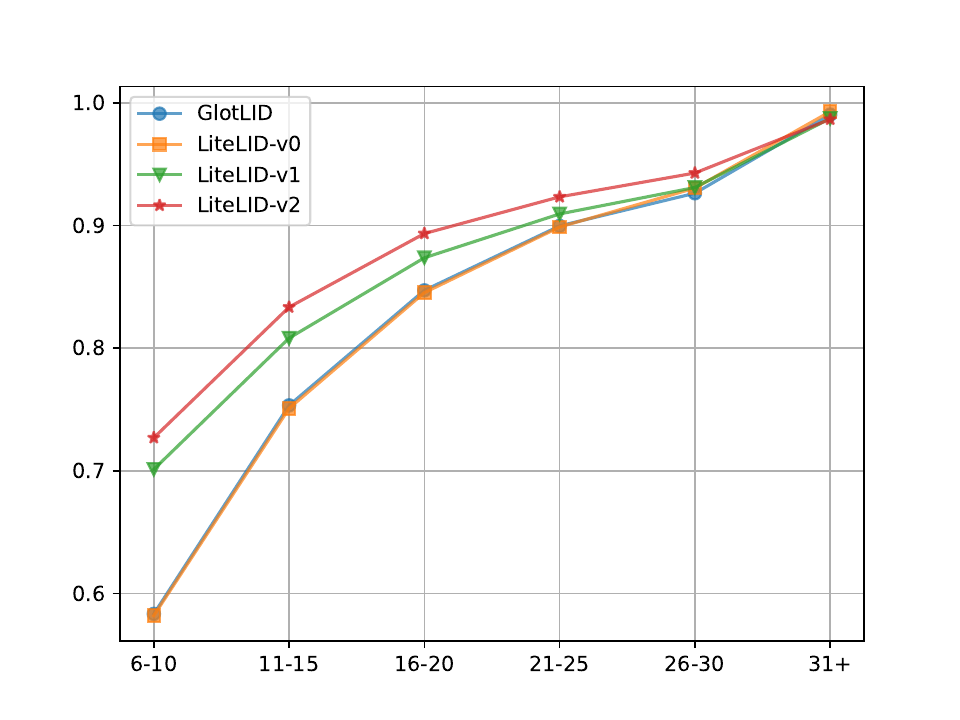}
    \caption{F1 as a function of the input length (in chars).}
    \label{fig:f1(nbchars)}
\end{figure}

As a sanity check, we also evaluated these models as pure LIDs, and observed that including isolated words in the training data was an effective way to improve the prediction for short segments, with no impact on longer ones. This is illustrated in Figure~\ref{fig:f1(nbchars)}: \glotlid{} and \wordlid-v0, which is identical to \glotlid{} trained with only 125~labels, are almost indistinguishable. \wordlid-v1, trained only on words, is also outperformed by \wordlid-v2, trained on sentences and words. \wordlid{}-v2 outperforms \glotlid{} on small segments, e.g., by 0.14 absolute points on sequences of length 6-10. \done\todo{Use consistent labels (GlotLID, etc) in the figure.}

\subsection{Improved \masklid{} results}

Equipped with this better backbone model, we ran experiments on our development data to compare their impact on \masklid{}. Results are in Table~\ref{tab:dev-results}. In the top part, we report baseline strategies for processing mixed-language data with a conventional LID: output the most likely language (k=1), the two most likely (k=2), or all languages whose probability exceeds a threshold (p>0.1). In all cases, we observe that \wordlid outperforms \glotlid{}, by a wide margin. When combined with \masklid{}, we also observe performance increases across the board when using an LID trained on the augmented dataset (compare \textbf{G} with \textbf{L} in Table~\ref{tab:dev-results}). These improved word-level probabilities yield clear improvements both for the code-switched and the monolingual test sets, for all values of the length parameter. 

\begin{table}[h]
    \centering
    \resizebox{\columnwidth}{!}{
        \begin{tabular}{ll|cccccc}
            \hline
                          & & \multicolumn{2}{c}{CS} & \multicolumn{2}{c}{mono} & \multicolumn{2}{c}{all}                             \\
            \textbf{Model} & config                 & EM                       & F1                      & EM   & F1   & EM   & F1   \\\hline
            \glotlid           & k=1                    & 0.0& 0.55& 0.98& 0.98& 0.49& 0.76 \\
            \glotlid           & k=2                    & 0.22& 0.56& 0.0& 0.66& 0.11& 0.61 \\
            \glotlid           & p>0.1                    & 0.09& 0.61&0.94& 0.97&0.51& 0.79\\
            \wordlid-v2         & k=1                    & 0.0& 0.63& 0.99& 0.99& 0.5& 0.81 \\
            \wordlid-v2         & k=2                    & \textbf{0.39}& 0.68& 0.0& 0.67& 0.19& 0.67 \\
            \wordlid-v2         & p>0.1                    & 0.18& 0.71&0.98& 0.99&0.58& 0.85 \\ \hline
            G+\masklid{}   & $\tau$=5              & 0.31 & 0.66 & 0.69 & 0.90 & 0.50 & 0.78 \\
            G+\masklid{}   & $\tau$=10              & 0.29& 0.66 & 0.74 &0.92 &0.51& 0.79 \\
            G+\masklid{}   & $\tau$=15              & 0.26 & 0.66 & 0.80 & 0.93 & 0.53 & 0.79 \\
            G+\masklid{}   & $\tau$=20              & 0.23& 0.65 &0.83 & 0.94& 0.53& 0.79 \\
           L-v2+\masklid{}   & $\tau$=5 (*)  &           \textbf{0.39}& \textbf{0.77}&0.96& 0.98&\textbf{0.68}& \textbf{0.88} \\
            L-v2+\masklid{}   & $\tau$=10             &0.35& 0.76 & 0.98& 0.99 &0.67& 0.87 \\
            L-v2+\masklid{}   & $\tau$=15               &0.30 &0.74 &0.99 &0.99 &0.65& 0.87 \\
            L-v2+\masklid{}   & $\tau$=20              & 0.25& 0.73 &0.99& 0.99 &0.62& 0.86 \\
        \end{tabular}}
    \caption{Exact match and F1 on the development data. (*) marks the best configuration.}
    \label{tab:dev-results}
\end{table}

For the next rounds of experiments, we use \wordlid{}-v2 as the backbone model, and set the length parameter $\tau$ to 5, as this is our best \masklid{} configuration on the development set.

\section{Computing better assignments with ILP \label{sec:ilpmodels}}
As discussed in \textsection{}\ref{ssec:masklid}, \masklid{} greedily assigns languages based on a global association score, computed as a sum of local scores, subject to several constraints. We propose to reformulate this problem in the ILP framework, enabling us to make the optimization objective explicit, to compute better solutions and to introduce additional constraints. 

\subsection{\genericname: The Core Program \label{ssec:coremodel}}
\done\todo{make this more compact}
Given a sequence $\sent$ of $T$ words, and a set of $L$ languages, we formalize CS LID as an optimal assignment problem, which maximizes under constraints a sum of word-language association scores, finally returning all languages assigned at least once. As an intermediate step, word-level labels are computed, which we model with a binary variable $y_{l,t} = 1$ if and only if $w_t$ is assigned to language $l$, 0 otherwise. We start with the following core program:
\begin{enumerate}
\item Input:
\begin{itemize}[noitemsep,topsep=0pt,parsep=0pt,partopsep=0pt]
\item $\sent = w_1 \dots w_T$ a sequence of words;
\item $\mathbf{C}(\sent) = \{c_{l,t}, l\in\interval{1}{L}, t \in \interval{1}{T}\}$\footnote{In the remainder, we drop the dependency on \sent{}.} the $L\times T$ matrix storing word-level language scores.\footnote{With \fasttext{}, these scores are logits computed in Eq.~\eqref{eq:matrix_v}.}\done\todo{positive or negative, higher is better}
\end{itemize}
    
\item Objective function, to be maximized w.r.t variables $\mathbf{Y} =\{y_{l,t}, l\in\interval{1}{L}, t \in \interval{1}{T}\}$:
\begin{equation}
  \mathcal{O}(\mathbf{Y})= \sum_{l=1}^{L}\sum_{t=1}^{T} c_{l,t} y_{l,t}. \label{eq:model1}
\end{equation}
\item Fundamental constraints:
    \begin{itemize}
        \item C1: at most 1 language per word
        \begin{align}
  \forall t, \sum_{l=1}^{L} y_{l,t} \le 1 \label{eq:m1:onelangperword}
\end{align}
        \item C2: at most $K$ languages per sentence
        \begin{align}
  \sum_{l=1}^L u_{l} \le K. \label{eq:m1:Klangpersent}
\end{align}
    \end{itemize}
\item Output:
\begin{itemize}[noitemsep,topsep=0pt,parsep=0pt,partopsep=0pt]
\item Binary assignments for variables in $\mathbf{Y}$, with $y_{l,t} = 1$ denoting the assignment of word $w_t$ to language $l$;
\item Binary assignments for variables in $\mathbf{U} = \{u_{l}, l \in \interval{1}{L}\}$, with $u_l=1$ if at least one word is assigned to language $l$. 
\end{itemize}
$\mathbf{U}$ and $\mathbf{Y}$ are linked through the following set of constraints (C0):
  \begin{equation}
  \forall l \in \interval{1}{L}: T\times u_{l} \geq \sum_{t=1}^{T} y_{l,t} \geq u_{l}. \label{eq:m1:nlanguages} 
\end{equation}
\end{enumerate}


Unconstrained optimization of objective~\eqref{eq:model1} yields a trivial solution where each word is assigned to all languages with a positive score: $y_{l,t}= \indic{c_{l,t} > 0}$.\footnote{$\indic{p}$ is the indicator function whose value is $1$ when predicate $p$ is \texttt{True}, $0$ otherwise.} Restricting words to be assigned to just one language (Eq.~\eqref{eq:m1:onelangperword}) fixes this issue, but yields another degenerate solution, where each word is mapped its most likely language: $y_{l,t}= \indic{c_{l,t} = \arg\max_{l'} c_{l',t} \wedge c_{l,t} > 0}$. 

\subsection{Improving \genericname{} }

We thus extend the core model in several ways to obtain more satisfactory solutions:
\begin{enumerate}
\item we explicitly rank assigned languages based on their global score.\footnote{This simulates the succession of language assignments performed in \masklid{}.} For this, we introduce a second index for variables in $\mathbf{U}$, with $u_{l,k}=1$ when $l$ has the $k^{th}$ largest global score. This implies to also introduce a third index to variables in $\mathbf{Y}$, yielding a generalized version of Eq.~\eqref{eq:m1:nlanguages}.
   \item we modify the objective function:
     \begin{itemize}
     \item by introducing a penalty $\mathbf{P}$ for every predicted language; 
     \item by introducing weights $\boldsymbol{\alpha}$ which constrain the ranking of language to reflect their respective global score;
     \end{itemize}
     The updated objective function thus becomes:
     \begin{align*}
       \mathcal{O}_{+}(\mathbf{Y}) =& \frac{1}{T}(\sum_{k=1}^K \alpha_k (\sum_{l=1}^{L}\sum_{t=1}^{T} c_{l,t} y_{k,l,t})) \\
                                                & -P\sum_{k=1}^K\sum_{l=1}^{L}u_{k,l}.  
     \end{align*}
    \item we finally consider supplementary constraints, each with its own meta-parameter:
    \begin{itemize}
        \item C3: sets a minimal length $\tau$ (in characters) required to identify a language, simulating the length constraint of \masklid{}.
        \item C4: a word $w_t$ can only be assigned to $l$ if $l$ is one of its $\alphapar$ most likely languages.
        \item C5: a word $w_t$ must be assigned to $l$ at rank 1 if $l$ is one of its $\alphapar$ most likely languages. 
        \item C6: the total number of language changes must be at most $S$.
    \end{itemize}
  \end{enumerate}
C4 and C5 are variants of the top-$\alphapar$ constraint of \masklid{}, while C6 is new and illustrates the flexibility of our framework.
A detailed formulation of the variables and constraints in \genericname\done\todo{fix name} are in Appendix~\ref{sec:formalization}.

\subsection{\genericname's best configurations}
We perform a search for the best settings for \genericname{} and identify three configurations that respectively maximize the EM on code-switched data (CS), on monolingual data (MONO), and on the average of the two (AVG). These reference configurations are as follows:
\begin{itemize}
\item CS: optimizes the core objective under constraints C1 and C2. 
\item MONO: optimizes the extended objective (with $\alpha_{1}=1, \alpha_2=0.7, P=20$), subject to C1, C2, C3 ($\tau=10$), C5 ($M=10$). 
\item AVG: optimizes the extended objective (with $\alpha_{1}=1, \alpha_2=0.75, P=15$), subject to C1, C2, C3 ($\tau=5$), C5 ($M=10$).
\end{itemize}
Performances on the development set using these configurations are in Table~\ref{sec:word-level-lid}, where AVG increases the \masklid{} baseline by 0.08 absolute points in EM (from 0.68 to 0.76). 
\begin{table}[h]
    \centering
    \resizebox{\columnwidth}{!}{
        \begin{tabular}{ll|cccccc}
            \hline
                           &               & \multicolumn{2}{c}{CS} & \multicolumn{2}{c}{mono} & \multicolumn{2}{c}{all}                                        \\
            \textbf{Model} & config        & EM                     & F1                       & EM                      & F1   & EM            & F1            \\\hline
            L-v2 + \masklid{}   & $\tau$=5  &           0.39& 0.77&0.96& 0.98&0.68& 0.88 \\
            L-v2 + \masklid{}   & $\tau$=10 & 0.35          & 0.76            & \textbf{0.98}                    & 0.99 & 0.67 & 0.87 \\
            L-v2 + \genericname & AVG           & 0.60                   & 0.83                     & 0.91                    & 0.97 & \textbf{0.76}          & 0.90          \\
            L-v2 + \genericname & CS            & \textbf{0.68}                   & 0.83                     & 0.06                    & 0.69 & 0.37          & 0.76          \\
            L-v2 + \genericname & MONO          & 0.47                   & 0.81                     & \textbf{0.98}                    & 0.99 & 0.73          & 0.90          \\
        \end{tabular}}
    \caption{Exact Match and F1 on the development data for various ILP configurations, compared to the \masklid{} baseline of Section~\ref{sec:word-level-lid}.
    }
    \label{tab:ilp-configs}
  \end{table}
  
 \section{Experiments on the testset}\label{sec:results}
 \subsection{Detecting language mixtures} 
 Using the models and parameter values identified on the development set, we finally process the test data and obtain the results in Table~\ref{tab:results-test}.
 
\begin{table*}[h]
  \centering

  \resizebox{0.9\textwidth}{!}{
 \begin{tabular}{clcccc|cccc|cccc}
        \\
        \hline
                        & & \multicolumn{4}{c}{\textbf{CS}} & \multicolumn{4}{c}{\textbf{mono}} & \multicolumn{4}{c}{\textbf{all}}                                                                      \\

        \textbf{Model}     & \textbf{config}                  & EM                                & F1                                & Pr   & Re   & EM   & F1   & Pr   & Re   & EM   & F1   & Pr   & Re   \\\hline
        \glotlid       & $k$=1                           & 0.0                               & 0.58                              & 0.87 & 0.44 & \textbf{0.97} & 0.97 & 0.97 & 0.97 & 0.49 & 0.78 & 0.92 & 0.71 \\
        \glotlid       & $k$=2                           & 0.20                              & 0.56                              & 0.56 & 0.56 & 0.0  & 0.66 & 0.49 & 0.98 & 0.10 & 0.61 & 0.53 & 0.77 \\
        \glotlid       & p>0.1                           & 0.06& 0.62& 0.85& 0.49&0.95& 0.97& 0.97& 0.98&0.5& 0.8& 0.91& 0.74 \\
        LiteLID-v2    & $k$=1                           & 0.0& 0.63& 0.95& 0.47&\textbf{0.99}& 0.99& 0.99& 0.99&0.50& 0.81& 0.97& 0.73 \\
        LiteLID-v2    & $k$=2                           & 0.36& 0.67& 0.67& 0.67&0.0& 0.67& 0.50& 1.0&0.18& 0.67& 0.58& 0.83 \\
        LiteLID-v2    & p>0.1                           & 0.15& 0.69& 0.91& 0.56&0.94& 0.98& 0.97& 1.0&0.54& 0.84& 0.94& 0.78 \\\hline \hline
         G+\masklid{}   &  $\tau$=5                      & 0.30& 0.68& 0.79& 0.60&0.76& 0.92& 0.87& 0.98&0.53& 0.80& 0.83& 0.79 \\
        G+\masklid{}   &  $\tau$=10                       & 0.29                              & 0.68                              & 0.82 & 0.59 & 0.80 & 0.93 & 0.89 & 0.98 & 0.54 & 0.81 & 0.85 & 0.79 \\
        L-v2+\masklid{} &  $\tau$=5                         &0.39& 0.78& 0.92& 0.67&0.94& 0.98& 0.97& 0.99&0.67& 0.88& 0.95& 0.83 \\
        L-v2+\masklid{} &  $\tau$=10                         &0.37& 0.77& 0.94& 0.66&\textbf{0.97}& 0.98& 0.98& 0.99&\textbf{0.67}& 0.88& 0.96& 0.83 \\\hline \hline
        L-v2+\genericname{} &  CS                      & \textbf{0.67}                              & 0.82                              & 0.83 & 0.82 & 0.04 & 0.68 & 0.52 & 1.00 & 0.36 & 0.75 & 0.68 & 0.91 \\
        L-v2+\genericname{}   & MONO & \textbf{0.42}                             & 0.79                              & 0.93                            & 0.69 & \textbf{0.96} & 0.98 & 0.98 & 0.99 & \textbf{0.69} & 0.89 & 0.96 & 0.84 \\
        L-v2+\genericname{} & AVG                      & \textbf{0.54}& 0.81& 0.88& 0.76&0.88& 0.96& 0.94& 0.99&\textbf{0.71}& 0.89& 0.91& 0.88 \\

    \end{tabular}

}

\caption{Results on the monolingual and CS test sets. $k$=1,2 when the LID always predicts $k$ languages; p>0.1 when it predicts any language having a probability >0.1. The three highest Exact Match values per column are in bold.}
    \label{tab:results-test}
\end{table*}\todo{SPECIFY CONFIGS; REPORT MORE RESULTS in Table~\ref{tab:results-test}}

They confirm, on a larger and more diverse set of languages, results observed on the development set: (a) as a bare model, \wordlid{}-v2 does better than \glotlid{} on all accounts, with very large gains on CS data; (b) changing the latter for the former when using \masklid{} also has large impacts, illustrated by a +0.13 absolute increase in EM; (c) changing the greedy optimizer for the ILP model yields additional gains, that are very significant for the CS subset. They amount, for instance, to a +0.17 absolute increase in EM for the AVG configuration, which also surpasses \masklid{} globally, albeit by a smaller margin (+0.04). Overall, the combination of our two improvements raises the average EM from 0.54 to 0.71 (AVG configuration). \done\todo{Masklid parameters}

McNemar's test confirms that the three ILP4LID configurations obtain significantly better EM scores than MaskLID on the test set: for $\alpha=0.05$ the $p$-values are respectively equal to $1.3\text{e-}302$ (AVG), $1.3\text{e-}18$ (MONO), and $0.0498$ (CS). The latter difference is borderline significant, which is hardly surprising as monolingual predictions for this configuration are quite inaccurate. 

Results broken down by language are in Table~\ref{tab:results-details}. They show improvements in the detection of CS text for all languages, with increases in EM up to +0.44 for Turkish-English or +0.35 for Nepali-English. These increases are partly due to the change of the backbone model, but also result from the improved optimizer (e.g., in Turkish-English and Turkish-German). Monolingual results are mostly affected by the change in the backbone LID, sharply improving the results of \masklid{}; \genericname{} does here slightly worse than \masklid{}, trading-off a small loss on monolingual data for large gains in CS test sets.    

\subsection{Error analysis}\label{ssec:error-analysis}

\paragraph{Comparing \genericname{} configurations} The CS configuration has a recall of 0.99 on monolingual data, but a very small Exact Match of 0.04. This is because using only C1 and C2 leads to predicting two languages in almost all cases, as explained in~\textsection\ref{ssec:coremodel}. For example, monolingual English sentences in this configuration are most often labeled with pairs involving a close language such as $\{$eng, nld$\}$ or  $\{$eng, fra$\}$; likewise, monolingual Spanish is often labeled as $\{$spa, por$\}$, etc. 

As for the MONO configuration, the most frequent mistake corresponds to only identifying one of the two correct languages in code-switched test sets: while the precision for this configuration is always high, with an average of 0.96, the recall for the CS data remains unsatisfactory (0.69). The AVG configuration is less prone to either kind of mistakes and represents our best trade-off between good CS detection abilities, and with very high scores in monolingual identification.

\paragraph{Comparing \genericname{} with \masklid} 
We focus now on the differences between \masklid{} and \genericname{}, when running with the same backbone model (\wordlid-v2). For this, we study the proportion of words that remain non assigned at the end of the process: in theory, assigning more words should lead to better global solutions.\footnote{Our objective functions can only increase when more words are assigned.} On the full test data, \masklid{} leaves 23\% of words without any language assignment - a sign that the greedy procedure makes early decisions that assign too many words to the first languages, leaving fragments that are too small to be assigned in a second round, thereby missing occurrences of CS text. By contrast, \genericname{} (AVG) is able to provide assignments for 97\% of the test words, yielding a clear increase of the recall for the CS test set (+0.1).

This error pattern is illustrated on following example of CS between \underline{French} and \textcolor{red}{Wolof}:
\begin{quote}
  \underline{parce que} \textcolor{red}{yooyu yépp ngir} \underline{vraiment} \textcolor{red}{wàññi} \underline{préssion} \textcolor{red}{picc yooyu ci ci ci toolu ceeb yi} \underline{égalment} \textcolor{red}{ak ci tool dugub yi}
\end{quote}
\masklid{} here correctly identifies all Wolof words, but also erroneously tags as Wolof ``vraiment'' and ``préssion", which are French words. It then fails to detect the presence of any French in the short remaining segment, leaving 14\% of words in the sentence unassigned. \genericname{} (AVG), by contrast, is here able to correctly identify both languages. 

\paragraph{Word-level assignments}
\label{sec:word-level-eval}
Even though this is not their primary goal, both \masklid{} and \genericname{} internally assign words to languages. Comparing how they perform this assignment also allows us to highlight the difference between them. Table~\ref{tab:wordlevel-eval} reports word-level performances on the code-switched test subsets containing reference word assignments: \masklid{} (\texttt{wordlid\_v2}, $\tau{=}5$) vs.\ \genericname{} in the CS configuration, macro-averaged over each pair's two languages, alongside the share of L1/L2 tokens each system leaves unassigned. Only tokens gold-labeled as L1/L2 are scored; an unassigned token is a false negative (lowering recall) but never a false positive. \masklid{}'s abstention maximizes its precision, having a clear overall negative effect on recall and F1. In comparison, \genericname{} has a slightly lower precision, but a much better recall. 
\begin{table}[t]
  \centering
  \resizebox{\columnwidth}{!}{
  \begin{tabular}{l*{8}{r}}
    \hline
    & \multicolumn{4}{c}{\textbf{ILP4LID (CS)}}  & \multicolumn{4}{c}{\textbf{MaskLID ($\tau{=}5$)}} \\
    \textbf{pair} & Pr & Re & F1 & Un & Pr & Re & F1 & Un \\
       \hline
    HIN-ENG            & 0.87          & 0.77              & 0.80          & 1.3         & 0.90 & 0.68 & 0.77       & 21.5 \\
    IND-ENG            & 0.94          & 0.90              & 0.92          & 0.4        & 0.92 & 0.82 & 0.87     & 8.3  \\
    NPI-ENG            & 0.88          & 0.70              & 0.77          & 3.3         & 0.94 & 0.57 & 0.71       & 32.5 \\
    SPA-ENG            & 0.93          & 0.83              & 0.88          & 0.9         & 0.95 & 0.75 & 0.84       & 15.7 \\
    TUR-DEU            & 0.98          & 0.95              & 0.96         &  0.5        & 0.99 & 0.84 & 0.91     & 13.5 \\
    TUR-ENG            & 0.92          & 0.91              & 0.91          & 0.3         & 0.94 & 0.71 & 0.80       & 18.6 \\

    \hline
    \textbf{macro avg} & 0.92 & 0.84 & 0.88  & 1.1 & 0.94 & 0.73 & 0.82 & 18.3 \\
    \hline
  \end{tabular}}
  \caption{Word-level evaluation results on code-switched test data. "Un" is the percentage of Unassigned words.}
  \label{tab:wordlevel-eval}
\end{table}

\paragraph{Remaining challenges} Very short, noisy, ambiguous social media sequences remain the biggest challenge. Even with our preliminary filtering, examples such as "auuuuuuuuush jajaja asi se habla lmao" can be found in the LinCE benchmark labeled as Spanish-English.
Our general recommendation is to use the AVG config, and CS/MONO if the number of languages in the data is known beforehand. 

\section{Related work}
\label{sec:related-work}

One way to detect CS is to develop \emph{word-level} LIDs. This problem is difficult, due to (a) extralexical terms (numbers or proper names), (b) lexical overlap (due to typological proximity or historical influences) between languages, (c)  occurrences of intra-word CS. This makes word-level LID intractable but for a very small number of languages. The bulk of studies along these lines have thus focused on simplifying settings where two \emph{predefined languages} are identified.
Supervised word-level LID is studied e.g. in \citep{nguyen-dogruoz-2013-word,king-abney-2013-labeling,elfardy-etal-2013-codeswitch,das-gamback-2014-identifying,al-badrashiny-diab-2016-lili, shehadi-wintner-2022-identifying}, where a rich mix of linguistic and orthographic features are used to detect code-switching points. \citet{sterner-2024-multilingual} considers more than two possible word classes (punctuations, named entities, mixed words) and detects CS between English and non-English
using Roberta-based \citep{conneau-etal-2020-unsupervised} classifiers.\done\todo{citation for Roberta} This system generalizes to (non-English) languages unseen in training. \citet{zhang-etal-2018-fast-compact} use a feed-forward network with global contstraints, identifying a fixed set about 100 language combinations selected for their relevance; most including English as one of the two languages. As far as we know, this tool has never been released. 

A first alternative framing considers \emph{infra-word LID}, predicting languages for characters \citep{kocmi-bojar-2017-lanidenn} or subwords \citep{mager-etal-2019-subword}, enabling the splitting of mixed-language words\footnote{Such cases appear for instance when a loan word in L2 needs to be inflected to match syntactic constraints in L1 - as in ``il a switché'' - (``\textsl{he switched}'') where ``switch'', an English borrowing, is conjugated in the past participle tense as a regular first group verb in French.}
and tagging each part with a distinct language. However, training subword models requires annotated data, which are not available on a large scale. Another approach studies \emph{unsupervised settings}, where the list of possible languages is not predefined. \citet{rijhwani-etal-2017-estimating} study unsupervised word-level LID to evaluate the amount of CS text on Twitter. \citet{kevers-2022-coswid} identify CS in multilingual documents, using a sliding window approach and relying on lists of language-specific words. A third alternative framing gives up on the prediction of word-level tags, generating a list of languages for each segment  \citep{stensby-etal-2010-language,lavergne-etal-2014-automatic}. \citet{kargaran-etal-2024-masklid} adopt this view and predict sentence-level labels, obtained by combining word-level scores computed by LIDs.

\done\todo{Move to related work or discussion}


\section{Conclusion and Outlook \label{sec:conclusion}}
In this work, we have proposed a reanalysis of the \masklid{} algorithm for CS LID. In early experiments, we identified a main issue with its strong reliance on word-level scores and mitigated it with a better backbone LID model, already outperforming the baseline system by a large margin. We then focused on the underlying optimization algorithm and developed an alternative version relying on the ILP framework, which offers a more flexible way to express constraints, and is also able to find better global solutions. This second change enabled us to further increase CS identification rates, yielding very large gains for some language pairs. These improvements constitute an important step towards a better processing of informal language styles, especially when they involve code-switching and/or minority languages.

In future work, we will continue exploring the tradeoffs between word-level and sentence-level predictions, training LIDs with better calibrated posterior probabilities that both \masklid{} and \genericname{} require, and also reporting confidence scores of our models. Another way to extend this work is the inclusion of more languages, especially languages usually written in non-Latin scripts but which can be transliterated in social media. Finally, optimizing the model to speed-up CS LID with the ILP model remains an important goal.  \done\todo{Improving LID improves masklid etc so is a sensible goal.}

\section*{Limitations}
\done\todo{write this}

\paragraph{Problem Framing} We frame the problem of CS LID as the identification of words associated to two distinct languages. We recognize that doing this is a double simplification. First, there are cases where more than two languages can occur in the same utterance as in the French-German-Luxemburgish dataset of \citet{lavergne-etal-2014-automatic}. Second, there are many cases where word-to-language assignments are ambiguous or ill-defined (punctuations, numbers, proper nouns, loan words), yielding uncertainty in the actual number of languages present in a sentence. As we use strict length limits for inserted L2 spans, our approach is bound to fail for very short segments, that are often difficult to unambiguously assign to just one language. For such highly ambiguous segments, a multi-label approach \citep{chifu-etal-2024-vardial,fedorova-etal-2025-multi} might be preferred. Finally, we note that our framing of CS LID also disregards \emph{intra-word} code-switching, where a borrowed base form in L2 is combined with a regular L1 inflection mark.

\paragraph{Language Coverage} Our experiments so far have only considered a handful of language pairs written in Latin script; they mostly cover informal speeches collected from social media or transcribed conversations. While our test is more diverse than previous work -- notably including several minority languages,  we recognize that it could be augmented with more language pairs and/or domains. While mixing languages that use different scripts should pose no major issue, it would still be interesting to include more data resulting from a romanization process. Our work with Hindi and Nepali is a first step in that direction, but is somewhat facilitated by the use of a regular romanisation process. Including CS data containing less standard forms of romanization, such as Arabizi for Arabic-French or Arabic-English \citep{seddah-etal-2020-building,shehadi-wintner-2022-identifying} would certainly prove to be more challenging, as high-quality LIDs for those types of mixed-language texts are difficult to find.

\paragraph{Computational Efficiency} The focus in this work has been primarily to better understand and diagnose the computations performed by \masklid{}, arguably the best available CS-LID model to date. For this, we have resorted to an ILP formulation, which can be seen as costly to run on personal computers. For instance, on a personal desktop (Intel Core i5 with 40GB RAM)\done\todo{Give processor description}, our CS LID detector takes about 13~s (for CS config) and 25~s (for AVG and MONO) to process 80 sentences of around 80 characters.\done\todo{fix numbers} This represents approximately x10 increase with respect to \masklid{} for short segments (<50), and a x20 increase of longer segments. \todo{Improve this with JZ; how does it compare with MaskLID ?} It should however be noted that our detector does not require any GPU and is distributable on multiple CPU cores.
\begin{figure}
    \centering
\includegraphics[width=0.49\linewidth]{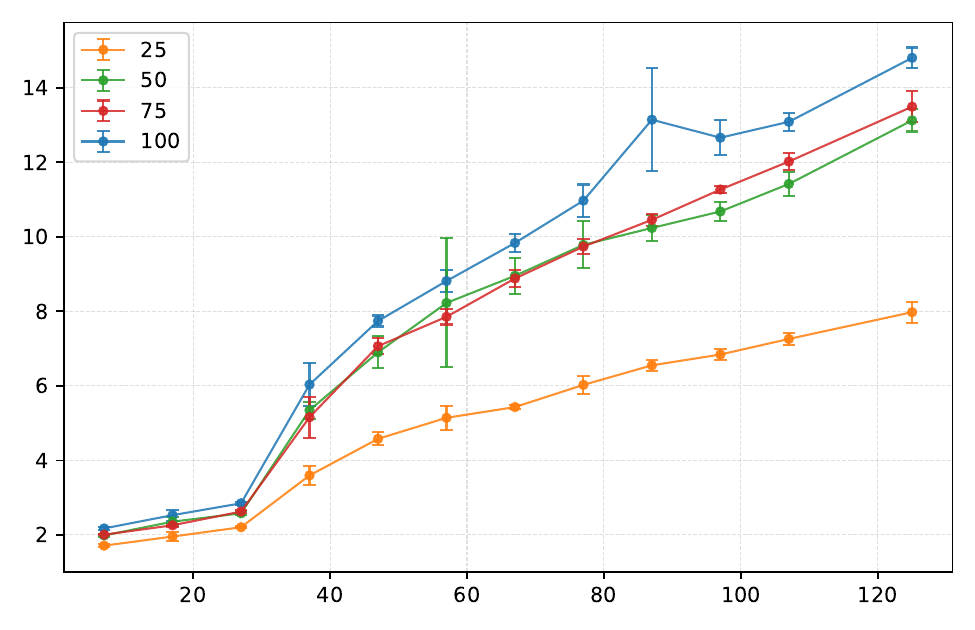}
\includegraphics[width=0.49\linewidth]{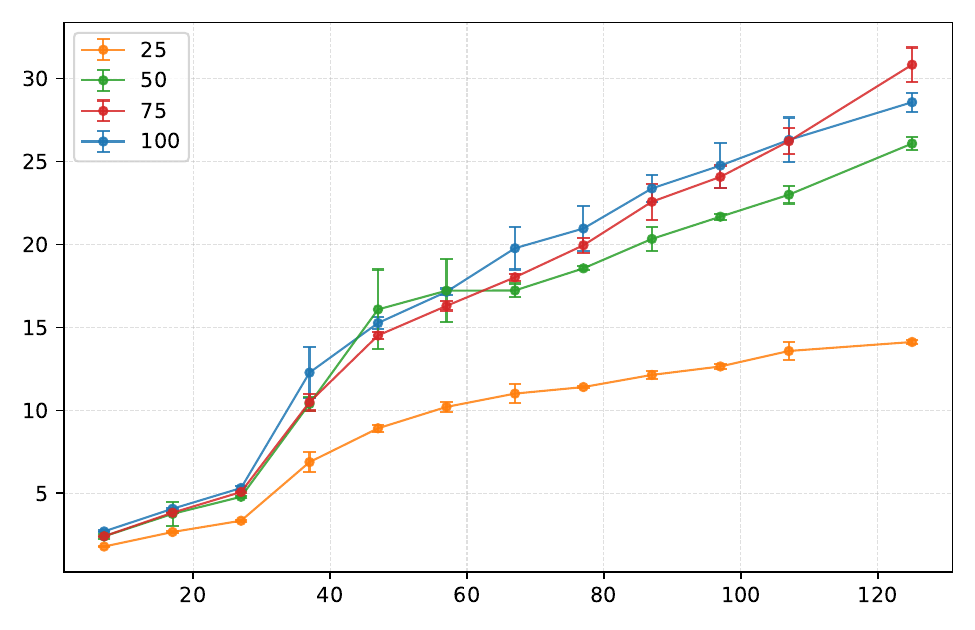}    
\caption{Runtime analysis showing a linear dependency between the number of language labels and the execution time on a desktop machine for 80 examples (in seconds). CS configuration (left) - AVG configuration (right). Curves correspond to buckets of different average lengths, ranging from \textasciitilde{}25 to \textasciitilde{}100 characters.}
\label{fig:runtimes}
\end{figure}
A straightforward way to improve this runtime would be to preselect the list of possible languages so as to reduce $L$, which directly defines the dimensions of the ILP program (see Appendix~\ref{sec:formalization}, Tables~\ref{tab:fullmodelone} and~\ref{tab:extendedmodel}).\done\todo{check section} This dependency is also illustrated on Figure~\ref{fig:runtimes}, which displays the variations of the ILP solver runtime with respect to $L$, for various typical sentence lengths.

For efficiency reasons, we have performed our experiments with Gurobi\footnote{\url{https://www.gurobi.com/}} version 13, a commercial solver, with free academic licences; our code can also be ran with any open-source tool supported by the Pyomo framework, e.g. HiGHS.\footnote{\url{https://highs.dev/}}

\section*{Acknowledgments} We thank Dwi Prima Handayani Putri for her contribution to early implementations of the ILP solver. We also thank Maxime Bouthors and Amir Hossein Kargaran for their inputs and feedback. We finally thank ARR reviewers and meta-reviewers for their constuctive comments. This work was performed using HPC resources from GENCI–IDRIS (Grant 2025-AD011017321).
François Yvon has been partly funded by the French National Funding Agency (ANR) under the France 2030 program (ref. ANR-23-IACL-0007).
\bibliography{anthology-1,custom}

\appendix

\section{\masklid: a Formal Account \label{sec:allmasklid}}
\done\todo{Formalize the complete algorithm}
A formal presentation of \masklid{} is in algorithm~\ref{alg:masklid}. It is based on \citep[Section~B]{kargaran-etal-2024-masklid} and an analysis of the accompanying Python source code.\footnote{\url{https://github.com/cisnlp/MaskLID}} The main difference with our informal presentation of \textsection\ref{ssec:masklid} is the second constraint that regulates the masking procedure. This constraint is controlled by three new parameters $M_{inc}, G_{ini}, G_{inc}$.

After computing the most likely language (line~\ref{line:mll}), two lists are computed, storing respectively the words with a strong (MaskLst) and weak (GangLst) association with $l^*$: they respectively contain all words having $l^*$ in their top-$M$ (respectively. top-$G$) most likely languages. Before masking the words in MaskLst (line~\ref{line:domask}), an additional constraint checks that \emph{based on the words in the GangLst}, $l^*$ reaches a predefined probability threshold (line~\ref{line:threshold}).\footnote{Precisely that $P(l^*|\operatorname{words}(GangList)) \ge P_{\min}$.} If this is not the case, the condition for inclusion in GangLst is weakened, yielding to the recruitment of more words in support of $l^*$. Critically, this also implies a weakening of the inclusion in the MaskLst, causing more words to be masked before searching for a possible second language. This means that \masklid{} incorporates a flexible mechanism to adapt the masking constraints depending on the global support of the optimal language for the current input segment. This mechanism, we contend, is key for the very good monolingual performance of \masklid.

\begin{algorithm}
  \caption{\masklid: Iterative Masking}
  \label{alg:masklid}
  $K$: Maximum number of iterations / languages \\
  $M_{\operatorname{ini}}, M_{\operatorname{inc}}$: Initial and increment values for {\bf Mask}  \\
  $G_{\operatorname{ini}}, G_{\operatorname{inc}}$: Initial and increment values for {\bf Gang}\\
  $\tau$: Minimum fragment length \\
  $P_{\min}$: Probability threshold \\
  $\sent = w_1 \dots w_T$
  \begin{algorithmic}[1]
    \State $k \leftarrow 1$
    \State $\operatorname{LangLst} \leftarrow \{\}$
    \While{$(len(\sent) > \tau \wedge k \le K)$}
    \State $l^* \leftarrow \arg\max_{l} P(l|\sent)$ \label{line:mll}
    \State $M \leftarrow M_{\operatorname{ini}},  G\leftarrow  G_{\operatorname{ini}}$
    \Repeat
    \State $\operatorname{MaskLst} \leftarrow \{\}$, $\operatorname{GangLst} \leftarrow \{\}$
    \For{$t= 1 \in [1:T]$}
    \If{$l^* \in \operatorname{TopLang}(w_t,M)$}
    \State $\operatorname{MaskLst}.\operatorname{insert}(w_t)$
    \EndIf
    \If{$l^* \in \operatorname{TopLang}(w_t,G)$}
    \State $\operatorname{GangLst}.\operatorname{insert}(w_t)$
    \EndIf
    \EndFor
    \State $M \leftarrow M + M_{\operatorname{inc}}$
    \State $G \leftarrow G + G_{\operatorname{inc}}$
    \State $P \leftarrow \max_{l} P(l|\operatorname{GangLst})$
    \Until{$(P \ge P_{\min})$} \label{line:threshold}
    \State $\sent \leftarrow \sent.\operatorname{remove}(\operatorname{MaskLst})$ \label{line:domask}
    \State $\operatorname{LangLst}.\operatorname{insert}(l^*)$
    \State $k \leftarrow k + 1$
    \EndWhile
    \State \Return $\operatorname{LangLst}$
  \end{algorithmic}
\end{algorithm}

This highlights an important difference with the way we implement the masking mechanism, as we rely on one single parameter, which remains constant accross sentences, input types, and languages. 

In \masklid, parameters $M_{\operatorname{inc}}$ and $G_{\operatorname{inc}}$ are set to $5$ and $P_{\min} = 0.9$. The recommended values for the other parameters are $M_{\operatorname{ini}}=3$, $G_{\operatorname{ini}} = 15$, $K=2$, $\tau = 20$ \citep[Section~C.2]{kargaran-etal-2024-masklid}. Except for the length parameter, that we also vary, these parameters are reused in all our experiments.\done\todo{fix this}

\section{Top-\alphapar{} lists are unreliable \label{sec:toplist-notrust}}\todo{Plot the same for wordlid}

\begin{figure*}[h!]
    \centering
        \includegraphics[width=0.49\linewidth]{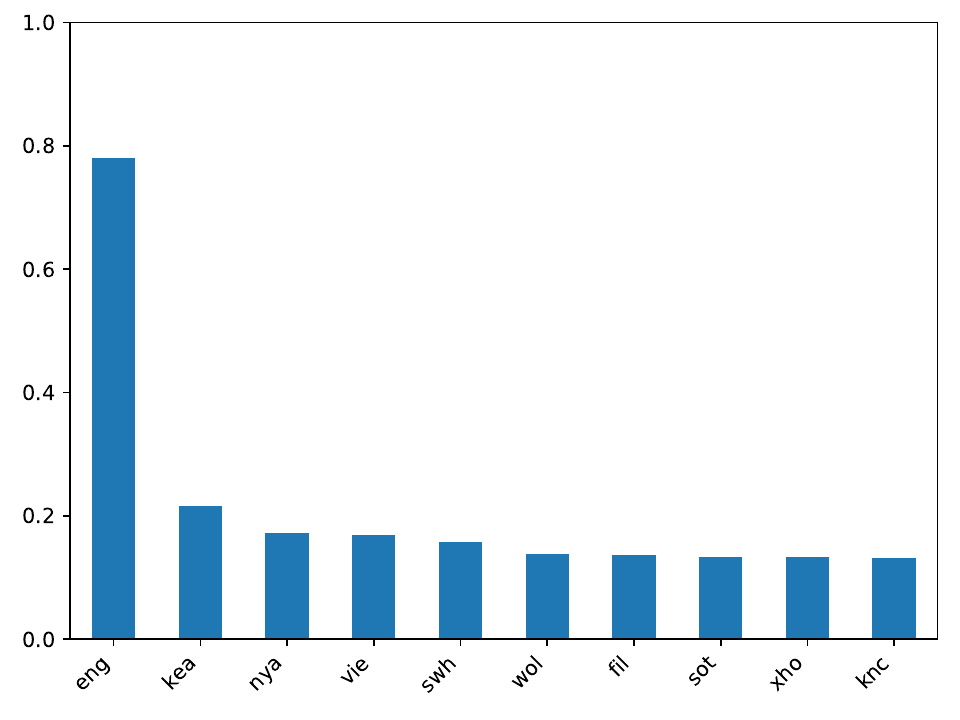}  
        \includegraphics[width=0.49\linewidth]{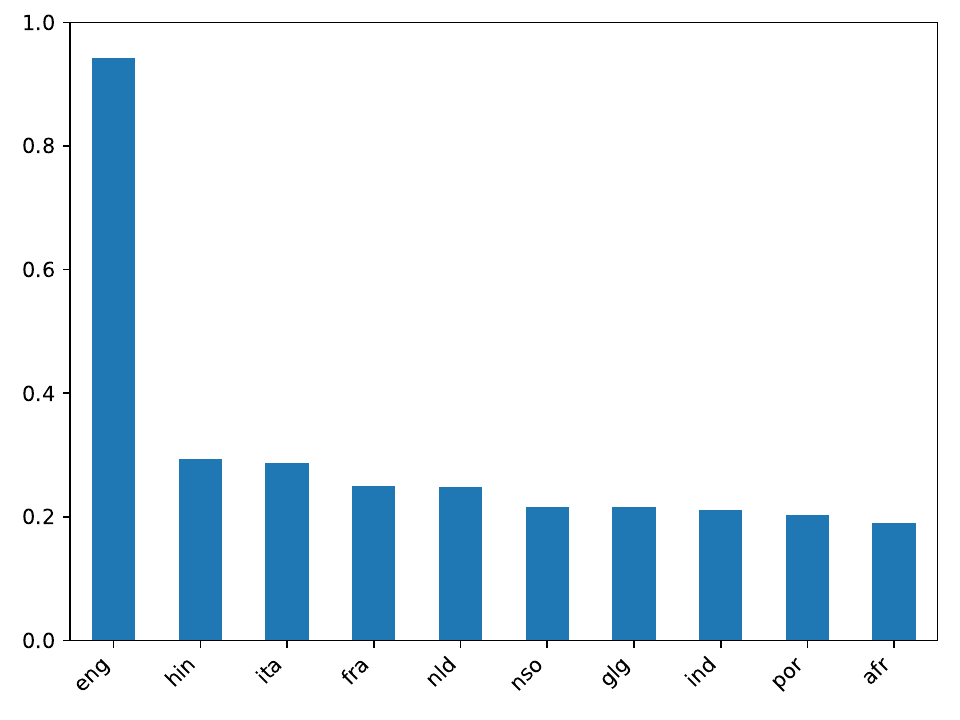}
        \includegraphics[width=0.49\linewidth]{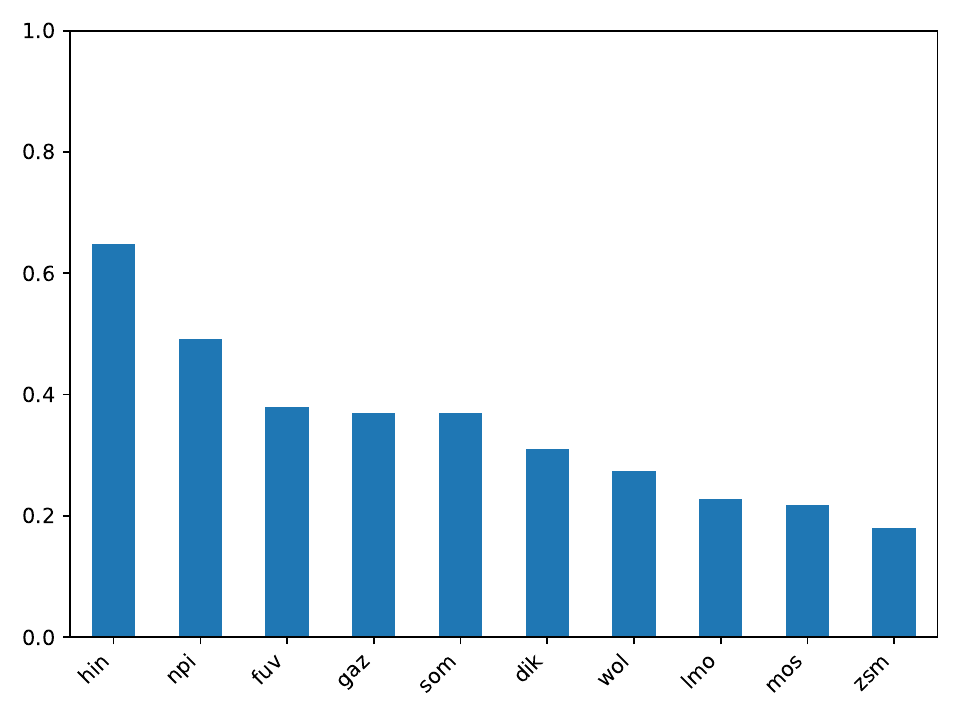}
      \includegraphics[width=0.49\linewidth]{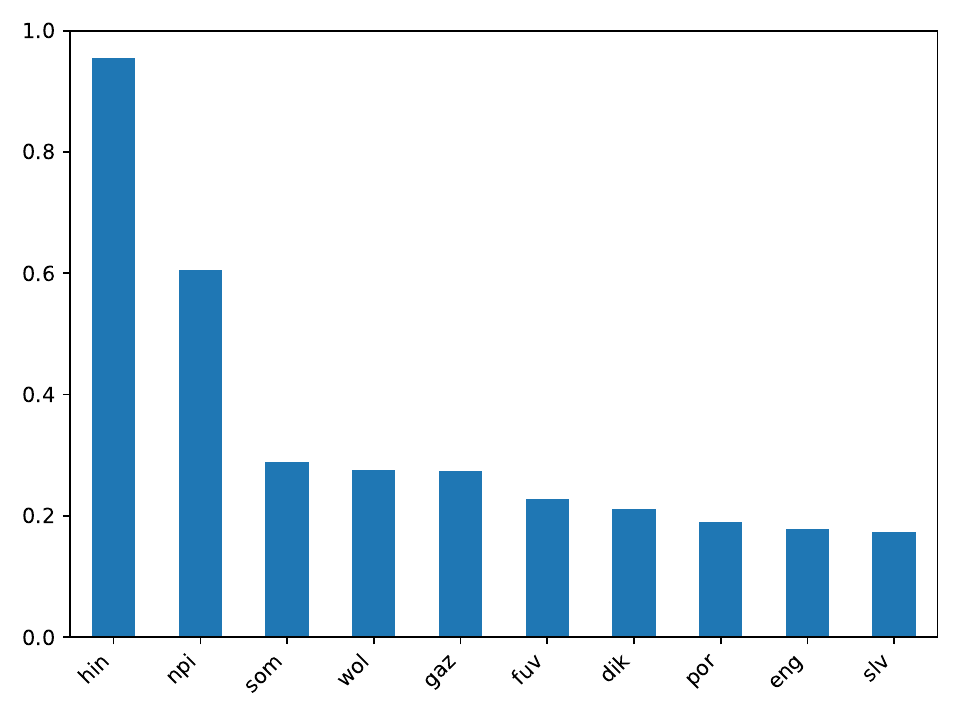}
    \caption{Percentage of cases when a given language label appears in top-10 languages for GlotLID (left) and \wordlid (right) in the monolingual development set. First row is English, second is Hindi. }
    \label{fig:enghintop10}
\end{figure*}

We argued in \textsection~\ref{ssec:issues-confidence} that \glotlid{} word-level scores were unreliable. We expand this analysis by looking at top-\alphapar{} lists. Figure~\ref{fig:enghintop10} reports the distribution of languages appearing in the top-10 languages for English and Hindi words in the corresponding monolingual development corpus. We observe the frequent occurrence of languages unrelated to English in the corresponding top-10 list (Kabuverdianu (kea), Wolof (wol), Nyanja (nya), Swahili (swh), Filipino (fil), etc). The same issue is visible for the list of Hindi words, with African languages such as Fulfude (fuv), Wolof (wol), Dinka (dik) well represented in the top-\alphapar{} list. \done\todo{how where those selected? FY: all dev monolingual}
Compared with our model, for both languages, the proportion of cases where the correct label is included increases (from 78 to 94 percent in English and from 65 to 96\% in Hindi). Additionally, languages typologically related to English (French, Dutch) start to appear.

\section{Languages \label{sec:labels}}

For this study, we focus on a restricted set of 125~languages written in Latin script, selected from the languages included in FLORES 200.\footnote{\url{https://huggingface.co/datasets/openlanguagedata/flores\_plus}} The list of labels is the following: (ISO~639-3 codes with Latin script): ace, afr, als, ast, ayr, azj, bam, ban, bem, bjn, bug, cat, ceb, ces, cjk, crh, cym, dan, deu, dik, dyu, ekk, eng, epo, eus, ewe, fao, fij, fil, fin, fon, fra, fur, fuv, gaz, gla, gle, glg, gug, hat, hau, hin, hun, ibo, ilo, ind, isl, ita, jav, kab, kac, kam, kbp, kea, kik, kin, kmb, kmr, knc, kng, lij, lim, lin, lit, lmo, ltg, ltz, lua, lug, luo, lus, lvs, min, mlt, mos, mri, nld, nno, nob, npi, nso, nus, nya, oci, pag, pap, plt, pol, por, quy, ron, run, sag, scn, slk, slv, smo, sna, som, sot, spa, srd, ssw, sun, swe, swh, szl, taq, tpi, tsn, tso, tuk, tum, tur, twi, umb, uzn, vec, vie, war, wol, xho, yor, zsm, zul.

This list covers a large variety of languages, covering multiple language families -- with the notable exception of languages not using the Latin script. For those languages, CS LID is a simple matter of script identification, which can be performed very reliably, e.g., using GlotScript \cite{kargaran-etal-2024-glotscript}.

\section{Training a Word-Level LID \label{sec:training-versions}}

\done\todo{training data, statistics, meta parameters, base performance as LID}
For training an alternative LID with more reliable word-level performances (\wordlid), our starting point (v0) are: the GlotLID-C corpus truncated to the 125 labels of interest, and the hyperparameters used for training GlotLID (v3). We use \fasttext{}'s Python API with $\texttt{lr}=0.8$, $\texttt{epochs}=1$, $\texttt{dim}=256$. We experiment with varying the hashmap size ($2e6$ seems fitting for our $125$~labels; we suspect that collisions in the hashmap - $1e6$ for over $2000$~labels - may explain some of GlotLID's  erroneous predictions) and the n-gram range. 

The crucial part is simple data augmentation. We split sentences into unigrams to create an augmented dataset, on the grounds that training on sentences \textit{and} words will make word-level scores more reliable. We source (up to) one million running words for each label.\footnote{For some languages in the corpus, the actual number of available words is less than 1M.} We tokenize by splitting on whitespaces. To mitigate the effect of the power-law distribution of words, we sub-sample words that occur more than 1000 times for each language, rejecting them with a 50\% chance.
Furthermore, we transliterate the Hindi and Nepali data that was in Devanagari in GlotLID-C, to improve recognition of these labels in Latin script.
We also experiment with enriching the training dataset with trigrams, but there were no clear benefits. In the main paper, we use a version trained on all sentences and up to 1 million words per label (v2).

Finally note that training a model only takes around 8 hours on a single CPU.
\begin{table}[!h]
\label{tab:fasttext_configs}
\centering
\resizebox{\columnwidth}{!}{
\begin{tabular}{lccl}
    \toprule
    \textbf{v} & \textbf{buckets} & \textbf{n-gram} & \textbf{data}                                   \\\midrule
    \textbf{v0}         & 1e6              & 2-5             & sents: GlotLID on 125 labels \\
    v0.1       & 2e6              & 2-5             & sents                                 \\
    v0.2       & 1e6              & 3-6             & sents        \\
    v0.3       & 2e6              & 3-6             & sents                          \\
    \textbf{v1}         & 2e6              & 3-6             & words                                      \\
    v1.1         & 1e6              & 2-5             & sents+words                                      \\
    v1.2       & 2e6              & 2-5             & sents+words                 \\
    v1.3       & 2e6              & 3-6             & sents+words \\

    v1.4         & 2e6              & 2-5             & sents+words+trigrams                \\
    v1.5       & 2e6              & 3-6             & sents+words+trigrams                   \\
    v1.6         & 2e6              & 2-5             & sents+words+romanized sents+trigrams   \\
        \textbf{v2}       & 2e6              & 2-5             & sents+words+romanized sents          \\
    \bottomrule
\end{tabular}}
\caption{FastText training configurations explored in training \wordlid.}
\end{table}

\section{\genericname{}: a Full Account \label{sec:formalization}}\done\todo{check names}

Integer Linear Programs are defined by a linear objective function of integer variables, and linear constraints (equalities or inequalities) involving the same set of variables. A recent introduction to ILP for NLP tasks is in \citep{srikumar-roth-2023-integer}.




\subsection{The core ILP model \label{ssec:ilp:core}}

Our ILP programs are defined by the following inputs:
\begin{itemize}[noitemsep,topsep=0pt,parsep=0pt,partopsep=0pt]
\item $L$ predicted languages $l \in [1:L]$
\item $\sent = w_1 \dots w_T$ a sequence of words;
\item $m_1 \dots m_T$ their respective lengths (in chars);
\item $\mathbf{C} = \{c_{l,t}, l \in [1:L], t \in [1:T]\}$ the matrix storing associations scores (in $\mathbb{R}^d$) between words and languages; 
\item $\mathbf{A} = \{a_{l,t}, l \in [1:L], t\in [1:T]\}$: $a_{l,t} = 1$ if $l \in \mathcal{L}_{\alphapar}(t)$, $0$ otherwise,
  where $\mathcal{L}_{\alphapar}(t)$ is the set of top-$\alphapar$ languages for $w_t$. 
\end{itemize}

The core model is defined over two sets of binary variables, storing respectively the word level and sentence-level language assignments:
\begin{itemize}[noitemsep,topsep=0pt,parsep=0pt,partopsep=0pt]
\item $\mathbf{Y}=\{y_{l,t}, l \in [1:L], t\in [1:T]\}$, with $y_{l,t} = 1$ denoting the assignment of word $w_t$ to language $l$;
\item $\mathbf{U} = \{u_{l}, l \in [1:L]\}$, with $u_l=1$ denoting that language $l$ globally assigned.
\end{itemize}

The objective function to be maximized is:
\begin{equation}
  \mathcal{O}(\mathbf{Y})= \sum_{l=1}^{L}\sum_{t=1}^{T} c_{l,t} y_{l,t}. \label{eq:coremodel}
\end{equation}
We also enforce the following sets of constraints:
\begin{itemize}
\item C0: dependency between $\mathbf{Y}$ and $\mathbf{U}$, imposing that $l$ is globally assigned ($u_l=1$) if at least one word is assigned to $l$: 
  \begin{align}
    \forall l, T\times u_{l} \geq \sum_{t=1}^{T} y_{l,t} \geq u_{l}. \label{eq:core:nlanguages} 
  \end{align}
\item C1: at most one language per word:
  \begin{align}
    \forall t, \sum_{l=1}^{L} y_{l,t} \le 1 \label{eq:core:onelangperword}
  \end{align}
\item C2: at most $K$ languages per sentence:
  \begin{align}
    \sum_{l=1}^L u_{l} \le K. \label{eq:core:Klangpersent}
  \end{align}
\end{itemize}

Figure~\ref{fig:example} illustrates these notations.\todo{fix the example as discussed}
{
\begin{figure*}[h]
\centering$\mathbf{w}$ = Identifying$_{eng}$ \textcolor{blue}{plusieurs}$_{fra}$ \textcolor{blue}{langues}$_{fra}$
\[
\begin{array}{ccc}

\mathbf{C(\sent)} =
\overset{\scriptstyle w_1\quad w_2\quad w_3}{
\begin{bmatrix}
c_{11}&c_{12}&c_{13}\\
\textcolor{blue}{c_{21}}&\textcolor{blue}{c_{22}}&\textcolor{blue}{c_{23}}\\
c_{31}&c_{32}&c_{33}
\end{bmatrix}
}
\mathbf{A} =\overset{\scriptstyle w_1\quad w_2\quad w_3}{
\begin{bmatrix}
1&0&0\\
1&1&1\\
0&1&1
\end{bmatrix}
}
\quad
\mathbf{Y} =
\overset{\scriptstyle w_1\quad w_2\quad w_3}{
\begin{bmatrix}
1&0&0\\
\textcolor{blue}{0}&\textcolor{blue}{1}&\textcolor{blue}{1}\\
0&0&0
\end{bmatrix}}
\quad
\mathbf{U} =
\begin{bmatrix}
1\\\textcolor{blue}{1}\\0
\end{bmatrix}
\begin{matrix}
\scriptstyle l_{1}\scriptstyle(\text{eng})\\\scriptstyle l_{2}\scriptstyle(\text{fra})\\\scriptstyle l_{3}\scriptstyle(\text{spa})
\end{matrix}

\end{array}
\]
\begin{equation*}
  \mathcal{O}(\mathbf{Y^*})= c_{11}+c_{22}+c_{23}
\end{equation*}
\caption{Code-switching LID as an ILP problem. The inputs are $\mathbf{C}(\sent)$, the cost matrix, and $\mathbf{A}$, which contains information about the presence of each language $l$ in the top-\alphapar{} list of each word.  $\mathbf{Y}$ is a feasible solution, with $\mathbf{U}$ the associated language vector. $\mathbf{Y}$ satisfies the constraints: at most one language for each word, no more than two languages ($\sum_l u_l =2$), a minimal length of $10$ chars for each span.
  $\mathcal{O}(\mathbf{Y^*})$ is the value of the objective function computed for this assignment.\label{fig:example}} 
\end{figure*}
}

Table~\ref{tab:fullmodelone} summarizes the variables and constraints in the core formulation. The number of variables grows with $T \times L$, while the number of constraints only grows in $T+L$; if $T$, the input length, cannot be changed, the number of languages in the underlying LID, could be easily adjusted on a per input basis, by only considering in the ILP languages that are sufficiently likely for at least one word.

\begin{table}
  \centering
  \begin{tabular}{ll}
    \hline
    \multicolumn{2}{l}{\textbf{Inputs}} \\ \hline
    $c_{l,t}$ & $T \times L$ (costs) \\
    $a_{l,t}$ & $T \times L$ (top-\alphapar{} languages)\\ \hline
    \multicolumn{2}{l}{\textbf{Variables}} \\ \hline
    $y_{l,t}$ & $T \times L$ (word assignments) \\
    $u_l$   & $L$ (sentence assignments) \\
    All & $(T + 1) \times L$\\         \hline
    \multicolumn{2}{l}{\textbf{Constraints}} \\ \hline
    Eq. \eqref{eq:core:nlanguages}    & $2 L$ (C0: count languages in \sent)\\
    Eq. \eqref{eq:core:onelangperword} & $T$ (C1: max. one language per word)\\
    Eq. \eqref{eq:core:Klangpersent} & 1 (C2: at most $K$ languages) \\
    All & $2L + T + 1$ \\ \hline
  \end{tabular}
  \caption{Counts of variables and constraints in the core model.}
  \label{tab:fullmodelone}
\end{table}
\done\todo{make a table with all variables and total size}

\subsection{The Extended Model \label{ssec:extended}}\done\todo{check name}

The core model differs from \masklid{} in one key aspect: it does not distinguish between dominant (L1) and embedded (L2) languages, which can cause the overprediction of CS texts. A typical error is when the ideal (monolingual) solution leaves some words unassigned; it may however happen that flipping some L1 words for L2 will increase the global objective (by assigning more words) and erroneously cause to predict two languages. In the following example, a Portuguese-based creole (Papiamento) is spuriously predicted in addition to English:
\begin{enumerate}[topsep=0pt,parsep=0pt,partopsep=0pt]
\item[]\textbf{ref}:  ``\sl i think i shoud stop going for swimming now'' (eng)
  
\textbf{pred}: `` \underline{\textcolor{red}{\sl i}} \sl think \underline{\textcolor{red}{\sl i shoud stop going for}} {\sl swimming now}''.
\normalfont The underlined span is assigned to pap (\textcolor{red}{\underline{Papiamento}}). \done\todo{Add a example ?}
\end{enumerate}

The extended ILP model explicitly enforces a hierarchy between languages, at the cost of an increase in the number of variables. Array $\mathbf{Y}$ is augmented with a third dimension which stores the rank of each language: rank~$1$ for the most likely language, rank~$K$ for the least likely.  The updated set of binary variables is: 
\begin{itemize}[topsep=0pt,parsep=0pt,partopsep=0pt]
  \item $\mathbf{Y}=  \{y_{k,l,t}, k\in\interval{1}{K}; l \in\interval{1}{L}, t\in\interval{1}{T}\}$ : $y_{k,l,t} =1$ if and only if $w_t$ is associated with $l$, which has rank $k$. 
  \item $\mathbf{U} = \{u_{k,l}, k \in\interval{1}{K}, l \in\interval{1}{L} \}$ with $u_{k,l} = 1$ if $l$ is the K$^{th}$ language. 
\end{itemize}

The extended objective function is defined as:
\begin{align}
  \mathcal{O}_{+}(\mathbf{Y}) =& \frac{1}{T}(\sum_{k=1}^K \alpha_k (\sum_{l=1}^{L}\sum_{t=1}^{T} c_{l,t} y_{k,l,t}))  \nonumber \\
                             & -P\sum_{k=1}^K\sum_{l=1}^{L}u_{k,l} \label{eq:extendedmodel},
\end{align}
and differs from objective in Eq.~\eqref{eq:coremodel} in two ways. It first contains parameters $1 \geq \alpha_1 > \dots > \alpha_K >0$, which ensure that the language with the largest global score will be at rank~1, as assigning it any other rank would decrease the objective value. In our implementation, we use $\alpha_1=1$ and $\alpha_2\in[0.25,0.99]$.\done\todo{values} Second, we introduce a second term as a fixed penalty $P$ for each new language. Tuning $P$ enables us to better control the trade-off between monolingual and bilingual assignments.

With these new variables, the fundamental constraints (C0-C2) take the following form:
\begin{itemize}
\item C0: dependency between $\mathbf{U}$ and $\mathbf{Y}$, imposing that $l$ is globally assigned at rank $k$ ($u_{k,l}=1$) if at least one word is assigned to $l$ at rank $k$: 
  \begin{align}
    \forall l,k:  T\times u_{k,l} \geq \sum_{t=1}^{T} y_{k,l,t} \geq u_{k,l}. \label{eq:extended:nlanguages} 
  \end{align}
\item C1: at most one language per word:
  \begin{align}
    \forall t, \sum_{k=1}^{K} \sum_{l=1}^{L} y_{l,t} \le 1 \label{eq:extended:onelangperword}
  \end{align}
\item C2: at most $K$ languages per sentence, at most one per rank \\
  \begin{align}
    \forall k,  \sum_{l=1}^L u_{k,l} \le 1 \label{eq:extended:onelangperrank} \\
    \sum_{l=1}^k \sum_{l=1}^L u_{k,l} \le K \label{eq:extended:Klangpersent} 
  \end{align}
\end{itemize}

We also consider additional constraints, each governed by its own meta-parameter:
\begin{itemize}
\item C3: sets a minimal length $\tau$ (in characters) required to identify a language, simulating the length constraint of \masklid{}; for this we introduce auxiliary variables in $\mathbf{Z}=\{z_{l,t}, l \in\interval{1}{L}, t\in\interval{1}{T} \}$:
  \begin{align}
    z_{l,t} - \sum_{k} y_{k,l,t} = 0  \nonumber \\
    \forall l, \sum_{t} m_t z_{l,t} \ge \tau \times (\sum_k u_{k,l}) \label{eq:extended:length}
  \end{align}
\item C4: $w_t$ can only be assigned to $l$ if $l$ is one of its $\alphapar$ most likely languages.\todo{check constraint and update the table accordingly}
  \begin{align}
    \label{eq:extended:topalphabis}
    \forall l,t, z_{l,t} + (1 - a_{l,t}) \le 1
  \end{align}
\item C5: $w_t$ must be assigned to $l$ at rank 1 if $l$ is one of its $\alphapar$ most likely languages.
  \begin{align}
    \forall l, t,  u_{1,l} + a_{l,t} + (1 - y_{1,l,t}) < 3 \label{eq:extended:topalpha}
  \end{align}
\item C6: the total number of language switches must be at most $S$:
  \begin{align}
    \sum_{l} \sum_{t>1} {s}_{l,t} \le & S         \label{eq:extended:maxswitch} \\
    \forall l, t,  {s}_{l,t} \ge &z_{l,t} - z_{l,t-1}            \nonumber     \\
    \forall l, t,  {s}_{l,t} \le &z_{l,t} - z_{l,t-1,} + (1 - r_{l,t}) \nonumber \\ 
    \forall l, t,  {s}_{l,t} \le & r_{l,t} \nonumber 
  \end{align}
  In this formulation, we introduce auxiliary variables $\mathbf{S}$ and $\mathbf{R}$ to keep track of language changes.
\end{itemize}

C3, C4 and C5 attempt to replicate constraints already existing in \masklid{}, with their corresponding meta-parameters. C6 is novel, and illustrates the flexibility of the ILP formalization. In practice, we found that relaxing C6 with a fixed penalty per language, as introduced in Eq.~\eqref{eq:extendedmodel}, yields better results, and is also more efficient.

\done\todo{make a table with all variables and total size}
\begin{table}[h]
  \centering
    \begin{tabular}{lp{0.7\columnwidth}}
    \hline
    \multicolumn{2}{l}{\textbf{Inputs}} \\ \hline
    $c_{l,t}$ & $T \times L$ (costs) \\
    $a_{l,t}$ & $T \times L$ (top-\alphapar{} languages)\\ \hline
    \multicolumn{2}{l}{\textbf{Variables}} \\ \hline
    $y_{k,l,t}$ & $K \times T \times L$ (word-level language and rank assignments) \\
    $u_{k,l}$   & $K \times L$ (sentence-level language and rank assignments)\\
    $z_{l,t}$ & $T \times L$ (word-level language assignment) \\
    $s_{l,t}$ & $T \times L$ (for Eq~\eqref{eq:extended:maxswitch})\\
    $r_{l,t}$ & $T \times L$ (for Eq~\eqref{eq:extended:maxswitch})\\ 
    All & $K \times T \times L + 3(T \times L) + (K \times L) $\\         \hline
      \multicolumn{2}{l}{\textbf{Constraints}} \\ \hline
      Eq. \eqref{eq:extended:nlanguages} & $2(K \times L)$ ( one lang. / per rank and word)\\
      Eq. \eqref{eq:extended:onelangperword} & $T$ ( one lang. / rank per word)\\
      Eq. \eqref{eq:extended:onelangperrank} & $K$ ( one lang. per rank)\\
      Eq. \eqref{eq:extended:Klangpersent} & 1 (at most $K$ languages) \\
      Eq. \eqref{eq:extended:length} & $L$ (min length / span)\\
      Eq. \eqref{eq:extended:topalphabis} & $T\times L$ (masking) \\
      Eq. \eqref{eq:extended:topalpha} & $T\times L$ (masking, alt. take) \\
      Eq. \eqref{eq:extended:maxswitch} & $1 + 3(T \times L)$ (switch. points)\\
    \textbf{All} & $2 (K\times L) + 4(T \times L) +$ \\ 
                 & $K + T +L + 1$ \\ \hline
  \end{tabular}
  \caption{Counts of variables and constraints in the extended model.}
  \label{tab:extendedmodel}
\end{table}
The variables and constraints in the extended model are in Table~\ref{tab:extendedmodel}. Here, the number of variables grows with $K \times T \times L$, which, given the typical values of $K$ (2 or 3), remains manageable and, compared to the core formulation, approximatly doubles the number of variables. The number of constraints grows linearly with $T$, $L$, and $K$, with the largest increase being caused by the use of C6. Finally, note that C4 and C5 can not be used simultaneously, so we use either one of the constraints or neither.


\section{Ablation of \genericname{} configurations}
\label{sec:ablation}

The impact of constraint selection, values of language weights and penalties, as well as constraint-specific hyperparameters $M, \tau, S$ is presented in Table \ref{tab:ablations}.
\begin{table}[h]
    \centering
    \resizebox{\columnwidth}{!}{
        \begin{tabular}{l|cccccc}
            \hline
                                                                      & \multicolumn{2}{c}{CS} & \multicolumn{2}{c}{mono} & \multicolumn{2}{c}{all}                      \\
            config                                                    & EM                     & F1                       & EM                      & F1   & EM   & F1   \\\hline

            \textbf{0}: CS                                                     &\textbf{0.68}                   & 0.83                     & 0.06                    & 0.69 & 0.37 & 0.76 \\
            \textbf{1}: 0+$\alpha_{k}=[1, 0.95]$                               & \textbf{0.68}                  & 0.83                     & 0.09                    & 0.70 & 0.39 & 0.77 \\
            1+$P=10$                                                  & 0.67                   & 0.84                     & 0.59                    & 0.88 & 0.63 & 0.86 \\
            1+$P=25$                                                  & 0.57                   & 0.83                     & 0.88                    & 0.96 & 0.73 & 0.90 \\
            1+$P=50$                                                  & 0.33                   & 0.76                     & \textbf{0.99}                    & 0.99 & 0.66 & 0.88 \\
            0+ $\boldsymbol{\alpha}=[1, 0.25]$                                 & 0.59                   & 0.80                     & 0.68                    & 0.91 & 0.64 & 0.86 \\
            0+ $\boldsymbol{\alpha}=[1, 0.50]$                                 & 0.65                   & 0.82                     & 0.47                    & 0.84 & 0.56 & 0.83 \\
            0+ $\boldsymbol{\alpha}=[1, 0.75]$                                 & 0.66                   & 0.82                     & 0.26                    & 0.77 & 0.46 & 0.80 \\
            0+ $\boldsymbol{\alpha}=[1, 0.99]$                                 & \textbf{0.68}                   & 0.83                     & 0.06                    & 0.69 & 0.37 & 0.76 \\
            1+C3 $\tau=5$                                             & \textbf{0.68}                   & 0.83                     & 0.17                    & 0.73 & 0.43 & 0.78 \\
            1+C3 $\tau=10$                                            & 0.65                   & 0.82                     & 0.36                    & 0.80 & 0.51 & 0.81 \\
            1+C3 $\tau=15$                                            & 0.61                   & 0.82                     & 0.49                    & 0.84 & 0.55 & 0.83 \\
            1+C5 $M=10$                                               & 0.66                   & 0.82                     & 0.46                    & 0.84 & 0.56 & 0.83 \\
            1+C5 $M=20$                                               & 0.63                   & 0.81                     & 0.57                    & 0.87 & 0.60 & 0.84 \\
            1+C5 $M=30$                                               & 0.62                   & 0.81                     & 0.65                    & 0.90 & 0.64 & 0.86 \\
            0+$\boldsymbol{\alpha}==[1, 0.95]$+C3,C5,P & 0.65                   & 0.82                     & 0.49                    & 0.84 & 0.57 & 0.83 \\
            0+$\boldsymbol{\alpha}==[1, 0.75]$+C3,C5,P & 0.64                   & 0.83                     & 0.82                    & 0.95 & 0.73 & 0.89 \\
            0+$\boldsymbol{\alpha}==[1, 0.50]$+C3,C5,P & 0.57                   & 0.83                    & 0.93                    & 0.98 & \textbf{0.75} & 0.91 \\
            AVG +$\boldsymbol{\alpha}=[1, 0.50]$                               & 0.46                   & 0.80                     & 0.97                    & 0.99 & 0.72 & 0.90 \\
            MONO+$\boldsymbol{\alpha}=[1, 0.50]$                               & 0.51                   & 0.82                     & 0.97                    & 0.99 & \textbf{0.74} & 0.91 \\
            MONO+$\boldsymbol{\alpha}=[1, 0.75]$                               & 0.35                   & 0.77                     & \textbf{0.99}                    & 0.99 & 0.67 & 0.88 \\
            1+C6 $S=3$                                                 & 0.67                   & 0.83                     & 0.09                    & 0.70 & 0.38 & 0.77 \\
            1+C6 $S=5$                                                 & 0.67                   & 0.83                     & 0.09                    & 0.70 & 0.38 & 0.77 \\
            \textbf{2}: AVG                                                       & 0.60                   & 0.83                     & 0.91                    & 0.97 & \textbf{0.76} & 0.90 \\
            \textbf{3}: MONO                                                      & 0.47                   & 0.81                     & \textbf{0.98}                    & 0.99 & 0.73 & 0.90 \\
        \end{tabular}}
    \caption{Exact match and F1 on the development data with various ILP configurations. "C3,C5,P" stands for C3 $\tau=5$, C5 $M=10$, $P=10$.}
    \label{tab:ablations}
\end{table}

Several points are worth pointing out. First, with a high enough language penalty, we can reliably reach a monolingual Exact Match that matches the recall of 0.99. The monolingual prediction errors stem mostly from wrongly identifying an additional language. However, this comes at the expense of identifying significantly fewer code-switching cases, which led us to choose a MONO configuration that had a monolingual EM of 0.98 instead of 0.99, as we deemed code-switching performance to be our principal goal.

Second, adding constraints C3 and C5 together with a handpicked penalty leads to a drastic increase in monolingual performance, at the expense of small drop in code-switched EM. 
Third, the model proves to be very sensitive to the value of language weights and penalty in the objective. 
We observe that the impact of the maximal number of switches constraint (C6) is limited. We have found that restricting number of switches has a minimally detrimental effect on CS EM, while having the disadvantage of doubling the runtime. 

C4 was also found to bring no significant improvement.

\begin{table*}[h!]
    \centering
    \begin{tabular}{lcccc|cccc|cccc}
                         & \multicolumn{4}{c}{\textbf{G + \masklid}} & \multicolumn{4}{c}{\textbf{L-v2 + \masklid}} & \multicolumn{4}{c}{\textbf{L-v2+\genericname{} AVG}}                                                                \\
                         & \multicolumn{4}{c}{(baseline)}           & \multicolumn{4}{c}{(first improvement)}    & \multicolumn{4}{c}{(second improvement)}                                                                       \\

                         & EM                                       & F1                                       & Pr                                            & Re   & EM   & F1   & Pr   & Re   & EM   & F1   & Pr   & Re   \\ \hline
        \hline
        eus-spa          & 0.59                                     & 0.86                                     & 0.94                                          & 0.79 & 0.55 & 0.87 & 0.99 & 0.77 & \textbf{0.73} & 0.9  & 0.95 & 0.86 \\
        hin-eng          & 0.1                                      & 0.56                                     & 0.7                                           & 0.47 & 0.17 & 0.68 & 0.9  & 0.55 & \textbf{\underline{0.34}} & 0.73 & 0.82 & 0.65 \\
        ind-eng          & 0.38                                     & 0.74                                     & 0.83                                          & 0.67 & 0.47 & 0.83 & 0.95 & 0.73 & \textbf{0.52} & 0.81 & 0.87 & 0.76 \\
        npi-eng          & 0.03                                     & 0.32                                     & 0.39                                          & 0.27 & 0.16 & 0.6  & 0.78 & 0.49 & \textbf{\underline{0.38}} & 0.68 & 0.71 & 0.65 \\
        spa-eng          & 0.1                                      & 0.63                                     & 0.83                                          & 0.51 & 0.18 & 0.68 & 0.88 & 0.55 & \textbf{\underline{0.36}} & 0.74 & 0.84 & 0.66 \\
        tur-deu          & 0.61                                     & 0.87                                     & 0.95                                          & 0.8  & 0.68 & 0.91 & 0.99 & 0.84 & \textbf{0.81} & 0.93 & 0.96 & 0.9  \\
        tur-eng          & 0.34                                     & 0.76                                     & 0.9                                           & 0.66 & 0.36 & 0.81 & 1.0  & 0.68 & \textbf{\underline{0.78}} & 0.92 & 0.96 & 0.89 \\
        wol-fra          & 0.16                                     & 0.71                                     & 0.93                                          & 0.58 & 0.35 & 0.8  & 0.99 & 0.67 & \textbf{\underline{0.37}} & 0.8  & 0.96 & 0.69 \\
        \hline
        Basque (eus)     & 0.93                                     & 0.98                                     & 0.96                                          & 1.0  & 0.99 & 0.99 & 0.99 & 1.0  & 0.93 & 0.98 & 0.96 & 1.0  \\
        English (eng)    & 0.92                                     & 0.97                                     & 0.95                                          & 0.99 & 0.99 & 0.99 & 0.99 & 1.0  & 0.95 & 0.99 & 0.98 & 1.0  \\
        French (fra)     & 0.92                                     & 0.98                                     & 0.96                                          & 1.0  & 0.96 & 0.99 & 0.98 & 1.0  & 0.95 & 0.98 & 0.97 & 1.0  \\
        German (deu)     & 0.84                                     & 0.96                                     & 0.92                                          & 1.0  & 0.97 & 0.99 & 0.99 & 1.0  & 0.91 & 0.97 & 0.95 & 1.0  \\
        Hindi (hin)      & 0.37                                     & 0.78                                     & 0.66                                          & 0.96 & 0.99 & 0.99 & 0.99 & 0.99 & 0.92 & 0.97 & 0.96 & 0.99 \\
        Indonesian (ind) & 0.89                                     & 0.97                                     & 0.94                                          & 1.0  & 0.96 & 0.97 & 0.97 & 0.98 & 0.88 & 0.96 & 0.93 & 0.99 \\
        Nepali (npi)     & 0.46                                     & 0.76                                     & 0.67                                          & 0.88 & 0.93 & 0.96 & 0.95 & 0.97 & 0.8  & 0.93 & 0.89 & 0.98 \\
        Spanish (spa)    & 0.94                                     & 0.97                                     & 0.96                                          & 0.99 & 0.97 & 0.98 & 0.98 & 0.99 & 0.87 & 0.96 & 0.93 & 0.99 \\
        Turkish (tur)    & 0.87                                     & 0.97                                     & 0.94                                          & 1.0  & 0.98 & 0.99 & 0.99 & 1.0  & 0.87 & 0.97 & 0.94 & 1.0  \\
        Wolof (wol)      & 0.82                                     & 0.95                                     & 0.91                                          & 1.0  & 0.94 & 0.98 & 0.97 & 1.0  & 0.77 & 0.94 & 0.88 & 1.0  \\
    \end{tabular}
    \caption{Detailed performances for monolingual and mixed language corpora. G stands for \glotlid, L for \wordlid-v2. Best performances for each code-switched language pair are in bold. Those for which the performance has been improved by a factor of two are underlined.}
    \label{tab:results-details}
\end{table*}

\section{Detailed Results per Language\label{sec:perlanguage}}
Table~\ref{tab:results-details} reports per language performance when using \glotlid{} and \wordlid{}-v2 as the base LIDs.

\todos{}

\end{document}